%% file: bare_jrnl_compsoc.tex
\makeatletter
\DeclareRobustCommand\onedot{\futurelet\@let@token\@onedot}
\def\@onedot{\ifx\@let@token.\else.\null\fi\xspace}

\def\eg{\emph{e.g}\onedot} 
\def\ie{\emph{i.e}\onedot}

\def\etal{\emph{et al}\onedot}
\makeatother

\documentclass[10pt,journal,compsoc]{IEEEtran}

\usepackage[linesnumbered,lined,vlined,ruled,commentsnumbered]{algorithm2e}

\usepackage{multirow} 
\usepackage{graphicx}
\usepackage{booktabs}

\usepackage{amssymb}
\usepackage{multirow,xspace}
\usepackage{makecell}
\usepackage{amsmath, eucal}

\ifCLASSOPTIONcompsoc
  \usepackage[nocompress]{cite}
\else
  \usepackage{cite}
\fi
\usepackage[pagebackref=false,breaklinks=true,letterpaper=true,colorlinks,bookmarks=false,citecolor={blue},linkcolor={blue}]{hyperref}

\usepackage{caption}
\usepackage[margin=4pt,font=footnotesize,labelfont=bf,labelsep=endash,tableposition=top]{caption}

\usepackage{subcaption}

\ifCLASSINFOpdf
\else
\fi

\begin{document}
\title{DeepEMD: Differentiable Earth Mover's Distance for Few-Shot Learning}

\author{Chi~Zhang, 
        ~
        Yujun~Cai,
        ~
        Guosheng~Lin,
        ~
        Chunhua~Shen% <-this % stops a space
\IEEEcompsocitemizethanks{
\IEEEcompsocthanksitem C. Zhang is with 
%the
School of Computer Science and Engineering, Nanyang Technological University, 
Singapore,
639798.
\protect\\
E-mail: chi007@e.ntu.edu.sg
\IEEEcompsocthanksitem 
    Y.  Cai is with % the
    Institute for Media Innovation, Nanyang Technological
	University, Singapore, 639798.
	    \protect\\
E-mail: yujun001@e.ntu.edu.sg

\IEEEcompsocthanksitem 
    G.  Lin is with %the
        School of Computer Science and Engineering, Nanyang Technological University, 
        Singapore, 639798.
        \protect\\
E-mail: gslin@ntu.edu.sg
\IEEEcompsocthanksitem 
    C.  Shen is with 
        % Monash University, Australia.
        Zhejiang University, China.
        \protect\\
E-mail: chunhua@me.com
        }
% <-this % stops an unwanted space
\thanks{
    Corresponding author: Guosheng Lin
}
}

\IEEEtitleabstractindextext{%

\input{1_abstract}

% \begin{abstract}
% In this paper, we address the few-shot classification task from 
% a new
% perspective of optimal matching between image regions. We adopt the Earth Mover's Distance (EMD) as a metric to compute a structural distance between dense image representations to determine image relevance. The EMD generates the optimal matching flows between structural elements that have the minimum matching cost, which is used to represent the image distance for classification.
% To generate the important weights of elements in the EMD formulation, we design a cross-reference mechanism, which 
% can effectively minimize the impact caused by the cluttered background and large intra-class appearance variations.
% To handle k-shot classification, we propose to learn a structured fully connected layer that can directly classify dense image representations with the EMD. Based on the implicit function theorem, the EMD can be inserted as a layer into the network for end-to-end training.
% We conduct comprehensive experiments to validate our algorithm and we set  new state-of-the-art performance on four popular few-shot classification benchmarks, namely miniImageNet, tieredImageNet, Fewshot-CIFAR100 (FC100) and Caltech-UCSD Birds-200-2011 (CUB).
% \end{abstract}

}

\maketitle

\IEEEdisplaynontitleabstractindextext

% \fi
\IEEEpeerreviewmaketitle

\input{2_introduction}

\input{3_related}

\input{4_0_preliminary}

\input{4_method}

\input{5_experiment}

\input{6_conclusion}

\ifCLASSOPTIONcaptionsoff
  \newpage
\fi

\bibliographystyle{IEEEtran}
\bibliography{egbib}



% \end{thebibliography}


% \end{IEEEbiography}

\begin{IEEEbiography}[{\includegraphics[width=1in,height=1.25in,clip]{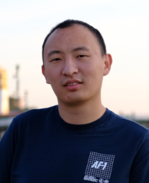}}]{Chi Zhang} is a PhD candidate with the School of Computer Science and Engineering, Nanyang Technological University, Singapore. He received the B.S. degree from China University of Mining and Technology in 2017.
His research interests are in computer vision and machine learning.
\end{IEEEbiography}
\vskip -1\baselineskip plus -1fil
\begin{IEEEbiography}[{\includegraphics[width=1in,height=1.35in,clip]{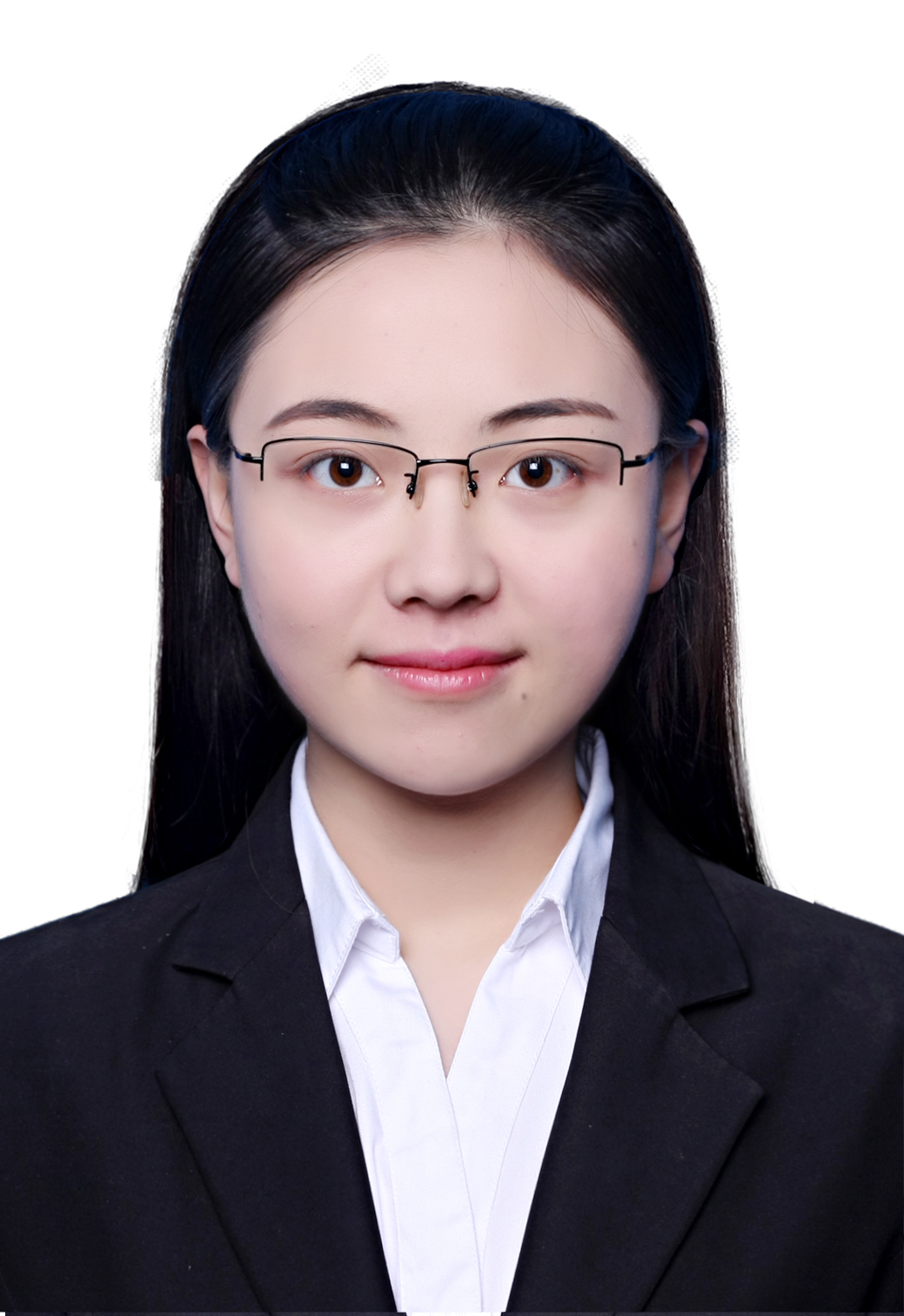}}]{Yujun Cai}
	received the B.Eng. degree in Information Science and Engineering from Southeast University in 2017. She is now a PhD candidate with the Institute for Media Innovation, Interdisciplinary Graduate School, Nanyang Technological University, Singapore. Her research interests mainly include computer vision, machine learning and human-computer interaction.
\end{IEEEbiography}
\vskip -1.0\baselineskip plus -1fil
\begin{IEEEbiography}[{\includegraphics[width=1in,height=1.25in,clip]{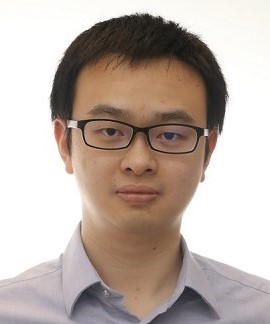}}]{Guosheng Lin} is an Assistant Professor at School of Computer Science and Engineering, Nanyang Technological University, Singapore. His research interests are in computer vision and machine learning.
\end{IEEEbiography}
\vskip -1\baselineskip plus -1fil
\begin{IEEEbiographynophoto}
{Chunhua Shen}
 is a Professor of Computer Science 
 at Zhejiang University, China.
\end{IEEEbiographynophoto}

%\newpage


%\vfill


\vfill

\end{document}

%% file: 1_abstract.tex
\begin{abstract}

% Deep learning has 
% proven 
% to be very effective in learning with a 
% large amount of labeled data. Few-shot learning in contrast attempts 
% to learn with only a few labeled data. 

In this work, we develop methods for
few-shot image classification 
from 
a new
perspective of optimal matching between image regions. We employ 
the Earth Mover's Distance (EMD) as a metric to compute a structural distance between dense image representations to determine image relevance. The EMD generates the optimal matching flows between structural elements that have the minimum matching cost, which is used to calculate 
the image distance for classification.
To generate the important weights of elements in the EMD formulation, we design a cross-reference mechanism, which 
can effectively 
alleviate 
the 
adverse 
impact caused by the cluttered background and large intra-class appearance variations.
To %handle
implement 
$k$-shot classification, we propose to learn a structured fully connected layer that can directly classify dense image representations with the 
EMD. Based on the implicit function theorem, the EMD can be inserted as a layer into the network for end-to-end training.
Our extensive 
experiments
validate the effectiveness of our algorithm which
outperforms
state-of-the-art 
methods
by a significant margin on five 
widely used 
few-shot classification benchmarks, namely, miniImageNet, tieredImageNet, Fewshot-CIFAR100 (FC100), Caltech-UCSD Birds-200-2011 (CUB), and CIFAR-FewShot (CIFAR-FS).
We also demonstrate the effectiveness of our method on the image retrieval task in our experiments. 
\end{abstract}

%% file: 2_introduction.tex
\IEEEraisesectionheading{\section{Introduction}\label{sec:introduction}}
\IEEEPARstart{D}{eep} neural networks have  
achieved 
great success in many vision tasks, typically requiring a large amount of labeled data. 
A notorious drawback of  deep learning methods is that they
suffer from poor sample efficiency. 
This is in sharp contrast to how we humans learn. 
In machine learning, few-shot  learning % is proposed  to
is the task that
addresses this issue,
which %can be seen 
is often solved  
as a special case of the broader meta learning.
Meta learning
attempts 
to learn a model that can generalize to new tasks with 
%minor
minimum 
adaption 
effort. 
One of the most well-studied test-beds for meta-learning algorithms is few-shot image classification, which aims to perform classification on new image categories with only a 
limited 
amount of labeled training data. This is the focus of %the
our 
work here.

To 
tackle this 
problem, a line of 
%the
previous 
work in literature adopts metric-based methods~\cite{matchnet,proto,feat,TADAM,relation,Revisiting} that learn to represent image data in an appropriate feature space and use a distance function to predict
image labels. Following the formulation of the standard image classification networks~\cite{cnn,resnet},
metric-based methods often employ a convolution neural network to learn image feature representations
and 
replace the fully connected layer with a distance function, 
\eg,
the 
\emph{cosine} distance and Euclidean distance. Such distance functions directly compute the distances between the embeddings of the test images and training images for classification, which bypasses the difficult optimization problem in learning a classifier in the few-shot setting. The network is usually trained by sampling from a distribution of tasks, in the hope of acquiring 
a good 
generalization ability to unseen but similar tasks.

\begin{figure}[t]
	\centering
	\includegraphics[width=1\linewidth]{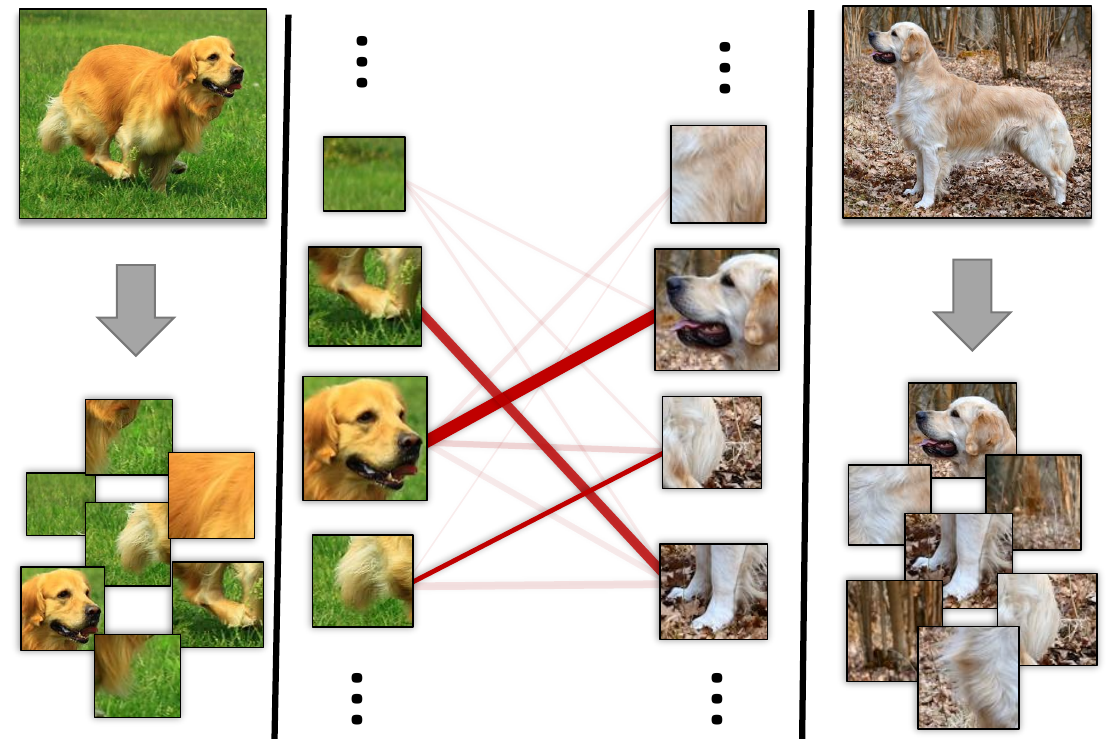}
	\caption{Illustration of using the Earth Mover's Distance for one-shot image classification. Our 
	method 
	uses an optimal matching cost between image regions to  represent the image distance more faithfully. 
	}
	\label{fig:introduction}
\end{figure}

Although these methods have achieved some degree of success, 
we observe that the cluttered background and large intra-class appearance variations may drive the image-level embeddings from the same category far apart in a given metric space. 
This issue 
can be largely alleviated by 
deep neural networks under the setting of fully supervised 
learning, 
thanks to the 
large capacity of deep models 
and 
sufficient 
training images. However,  it is almost inevitably amplified in low-data regimes and thus
adversely 
impacts the image classification accuracy. Moreover, a mixed global representation 
would struggle to well capture 
image structures and 
is likely to 
lose useful local feature characteristics.
Local features can provide discriminative and transferable information across categories, which can be important cues for image classification in the few-shot scenario. Therefore, a desirable metric-based algorithm should have the ability to 
exploit the local discriminative information and  minimize the 
distraction 
caused by irrelevant regions.

A natural 
approach 
to determine the similarity of two complex structured representations is to compare their building blocks.  The difficulty lies in that we do not have their correspondence supervision for training and not all building elements 
can always find their counterparts in the other structures. To solve the problems above, in this paper, we formalize the few-shot classification as an instance of optimal matching, and we propose to use the optimal matching cost between two structures to represent their dissimilarity.  Given the local feature representation sets generated by two images, we 
use 
the Earth Mover's Distance (EMD)~\cite{emd} to compute their structural similarity. The EMD is 
a
metric for computing distance between structured representations, which was originally proposed for image retrieval. Given the distance between all element pairs, the EMD can acquire the optimal matching flows between two structures that have the minimum overall distance. It can also be interpreted as the minimum cost to reconstruct a structured representation 
against 
the other one. 
An illustration of our motivation is shown in Fig.~\ref{fig:introduction}. The EMD has the formulation of the transportation problem~\cite{transportation} and the global minimum can be 
attained 
by solving a Linear Programming problem. To embed the optimization problem into the model for end-to-end training, we
apply the implicit function theorem~\cite{krantz2012implicit,dontchev2009implicit,barratt2018differentiability} to form the Jacobian matrix of the optimal optimization variables with respect to the problem parameters~\cite{barratt2018differentiability}. We explore multiple ways to extract local representations from an image, including fully convolutional networks, image grids, and image region sampling. We also investigate pyramid structures at both the feature level and the image level to capture local representations at different scales.

An important problem-specific 
parameter in the EMD formulation is the weight of each element. Elements with large weights generate more matching flows and thus contribute more to the overall distance. Ideally, the algorithm should 
accommodate 
the flexibility to assign less weight to irrelevant regions such that they contribute less to the overall distance no matter which elements they match with. To achieve this goal, we propose a cross-reference mechanism to determine the importance of the elements. In 
the proposed 
cross-reference mechanism, we determine the weight of each node by comparing it with the global statistics of the other structure. 
Intuitively, the image region that shows greater  relevance to the other image is more likely to be the object region and should be assigned with % more 
a larger 
weight, while the weights of  high-variance background regions and the object parts that are not  co-occurrent in two images should be eliminated as much as possible %for
when 
computing the  matching cost.

In the $k$-shot setting where multiple support images are presented, we propose to learn a structured fully connected (FC) layer as the classifier for classification to make use of the increasing number of training images. The structured FC layer includes a group of learnable vectors for each class.
At inference time, we use the EMD to compute the distance between the image embeddings and the learnable vector set in each class for classification.
The structured FC is an extension of the standard fully connected layer in that it replaces dot product operations between vectors with 
the 
EMD function between vector sets such that the structured FC layer
can 
directly classify feature maps. The structured FC layer can also be interpreted as learning the prototype embeddings generated by a dummy image for each category such that the test images can be matched with each of them for classification.

To validate our algorithm, we conduct extensive experiments on multiple datasets to demonstrate the effectiveness of our algorithm. Our main contributions are summarized as follows:
\begin{itemize}

	\item We propose to formalize the few-shot image classification as an optimal matching problem and %adopt
	employ
	the Earth Mover's Distance as the distance metric between structured representations. The EMD layer can be embedded into the network for end-to-end training.
	
	\item We propose a cross-reference mechanism to generate the weights of elements in the EMD formulation, which can effectively reduce the noise introduced by the irrelevant background regions in 
	% the 
	images.
    
    \item We propose to learn a structured fully connected layer in the $k$-shot settings, which 
    %could
    is able to 
    directly classify the structured representations of  an image using the Earth Mover's Distance.
    
    \item Experiments on five popular few-shot classification benchmark datasets---miniImagenet, tieredImagenet, FC100, CUB, and CIFAR-FS show that our algorithm on both 1-shot and 5-shot classification tasks significantly outperforms the baseline methods and achieves new state-of-the-art performance. We also demonstrate that our method can effectively improve many deep metric learning methods on the image retrieval task. 

\end{itemize}

Our preliminary result 
%is
was 
published in~\cite{zhang2020deepemd}. To facilitate future research, the source code and trained
models
% will be 
are made 
available at this link\footnote{%https://github.com/icoz69/DeepEMD
%
%
% CS: This link alias is simpler 
%
https://git.io/DeepEMD
}.

%% file: 3_related.tex
\section{Related Work}
\textbf{Few-Shot Learning.} 
The research literature on few-shot learning  shows great diversity
\cite{Keshari18,FeiFeiFP06,GidarisCVPR2018,YanZH19,Wertheimer_2019_CVPR,wang2020instance,afrasiyabi2020associative,liu2020negative,lichtenstein2020tafssl,yu2020transmatch,xuattentional}.
There are two main streams in the few-shot classification literature, metric-based approaches and optimization-based approaches. 
Optimization-based methods, 
\eg, 
%\cite{maml,MunkhdalaiICML18,trainmaml,SantoroBBWL16,closer,MunkhdalaiICML2017,ScottNIPS2018,PPA,Zhou2018deep,yaoyao,MetzICLR19,LeeICML18,LiICML2018,Luketina2016,Naik92,Jamal_2019_CVPR,dense,Schonfeld_2019_CVPR,flennerhag2019meta,li2019learning,park2019meta,franceschi2018bilevel,rajeswaran2019meta} 
\cite{MunkhdalaiICML18,SantoroBBWL16,closer,MunkhdalaiICML2017,ScottNIPS2018,PPA,Zhou2018deep,MetzICLR19,LeeICML18,LiICML2018,Luketina2016,Naik92,dense,Schonfeld_2019_CVPR,flennerhag2019meta,li2019learning,park2019meta,franceschi2018bilevel,rajeswaran2019meta,FranceschiICML18,qi2018low,simon2020adaptive,kim2020model,zhangiept,fei2021melr,oh2021boil,snell2020bayesian,patacchiola2020bayesian}, 
target at effectively adapting model parameters to new tasks in the low-shot regime. 
For example, MAML~\cite{maml} and many of its variants~\cite{trainmaml,leo,yaoyao,Jamal_2019_CVPR} aim to learn a good model initialization that can rapidly adapt to novel tasks with limited optimization steps.
Our design is more related to the metric-based methods~\cite{matchnet,proto,feat,TADAM,relation,Revisiting,CAN,ctm,tieredimagenet,liu2019prototype,xing2019adaptive,Zhang_2021_CVPR}, which aim to represent samples in an appropriate feature space where data from different categories can be distinguished with distance metrics. 
To achieve this goal,
most previous methods represent the whole image as a data point in the feature space. There are also some works utilizing local features to make predictions. For example, Lifchitz~\etal~\cite{dense} directly make predictions with each local feature and fuse their results. Li \etal~\cite{Revisiting} adopt $k$-NN to fuse local distances. Cross Attention Networks~\cite{CAN} use attention mechanisms to highlight the target object regions and generate  discriminative features for few-shot classification. 
CrossTransformers~\cite{crosstrans} compute the distances between  spatially-corresponding features in the query and labeled samples for few-shot classifications.
Our solution to the $k$-shot problem also draws connections to optimization-based methods since  we learn a classifier that can directly classify structured representations with the Earth Mover's Distance, which can benefit from the increasing number of support samples.

Besides the two popular approaches, many other promising methods have also been proposed to tackle the few-shot classification problem, such as works based on graph theories~\cite{Kim_2019_CVPR,wDAE,graph,yang2020dpgn}, reinforcement learning~\cite{rl}, differentiable SVM~\cite{metaoptnet}, generative models~\cite{Generative,Imaginary,SchwartzNIPS18,Mehrotra2017,ZhangNIPS2018MetaGAN,Hallucination,shen2019learning,li2020adversarial,yang2021free}, transductive learning~\cite{rodriguez2020embedding,dhillon2019baseline,hu2020empirical,yue2020interventional,boudiaf2020transductive}, recurrent models~\cite{shyam2017attentive,lstmmeta}, self-supervised learning~\cite{su2020does,tian2020rethinking}, 
the recent capsule network~\cite{wu2020attentive}, and temporal convolutions~\cite{SNAIL}.
Few-shot learning has  also been investigated for other computer vision tasks, such as image segmentation~\cite{zhang2019canet,zhang2019pyramid,liu2020crnet} and object detection~\cite{yang2020contexttransformer}.

\textbf{Earth Mover's Distance.}  Earth Mover's Distance (EMD) was originally proposed in~\cite{emd} as a metric for color and texture based image retrieval.  EMD has the formulation of the well studied transportation problem 
%from linear optimization,
in linear programming, 
and thus the global optimal matching can be found by solving a 
% Linear Programming problem. 
linear program.
EMD has several desirable properties that make it a popular method to compare structured representations. First, EMD can generate a structural similarity without explicit alignment information. It extends the distance between single elements to the distance between sets or distributions. Second, the number of elements in the sets can vary and EMD allows for partial matching when the total weights of two sets are not equal.
EMD has been widely applied to many areas. For example, Kusner \etal\ \cite{kusner2015word} use EMD to measure the similarity between two documents, which calculates the minimal cost to transfer the word embeddings in a document to the other  for document classification.
Wang and Chan~\cite{wang2014new} propose to represent the hand shapes and textures with superpixels and employ EMD to measure the dissimilarity between the hand gestures for gesture recognition. 
In~\cite{nikolentzos2017matching}, Nikolentzos \etal\  represent graph data  as a set of vectors corresponding to the vertices and use EMD to determine the similarity of two graphs for graph comparison. Schulter \etal~\cite{schulter2017deep} solve the multi-object tracking problem with a network flow formulation that learns features for network-flow-based data association.
Zhao \etal~\cite{zhao2008differential} propose to use the differential EMD to % handle
tackle the 
visual tracking problem based on the sensitivity analysis of the simplex method.
Li \cite{li2013tensor} uses a tensor-SIFT based EMD to tackle the contour tracking
problem.

\textbf{Parameterized optimization.}
Parameterized optimization problems have  a parameterized  object function and constraints that depend on input data. 
Many previous works have investigated differentiation through the $\rm argmin$ operators.  In \cite{gould2016differentiating}, Gould \etal present the methods for differentiation through optimization problems with only equality constraints.
Agrawal \etal \cite{agrawal2019differentiating} propose a method that can compute the gradient of the solution with respect to the coefficients in the convex cone program, which can scale to large problems.
Barratt \cite{barratt2018differentiability} describes the general case of  using the implicit function theorem and interior point methods to compute the Jacobian of the solution with respect to the problem parameters.
With the same theory, Amos and Kolter \cite{amos2017optnet} design a batched Quadratic Programming solver as a layer that can be integrated into a neural network for end-to-end training. 
In~\cite{agrawal2019differentiable},  Agrawal \etal\  propose a differentiable convex optimization layer that can differentiate through disciplined convex programs and allow users to define problems in a natural syntax without converting problems to canonical forms. 
Vlastelica \etal\ \cite{vlastelica2019differentiation} introduce
the combinatorial building blocks into neural networks and the end-to-end trainable network can generate informative backward gradients through any black-box implementations of combinatorial solvers. Based on such building blocks, Rolínek~\etal~\cite{rolinek2020deep} design an end-to-end network that incorporates a combinatorial solver to solve the graph matching problem.

\begin{figure*}[]
	\centering
	\includegraphics[width=0.99\linewidth]{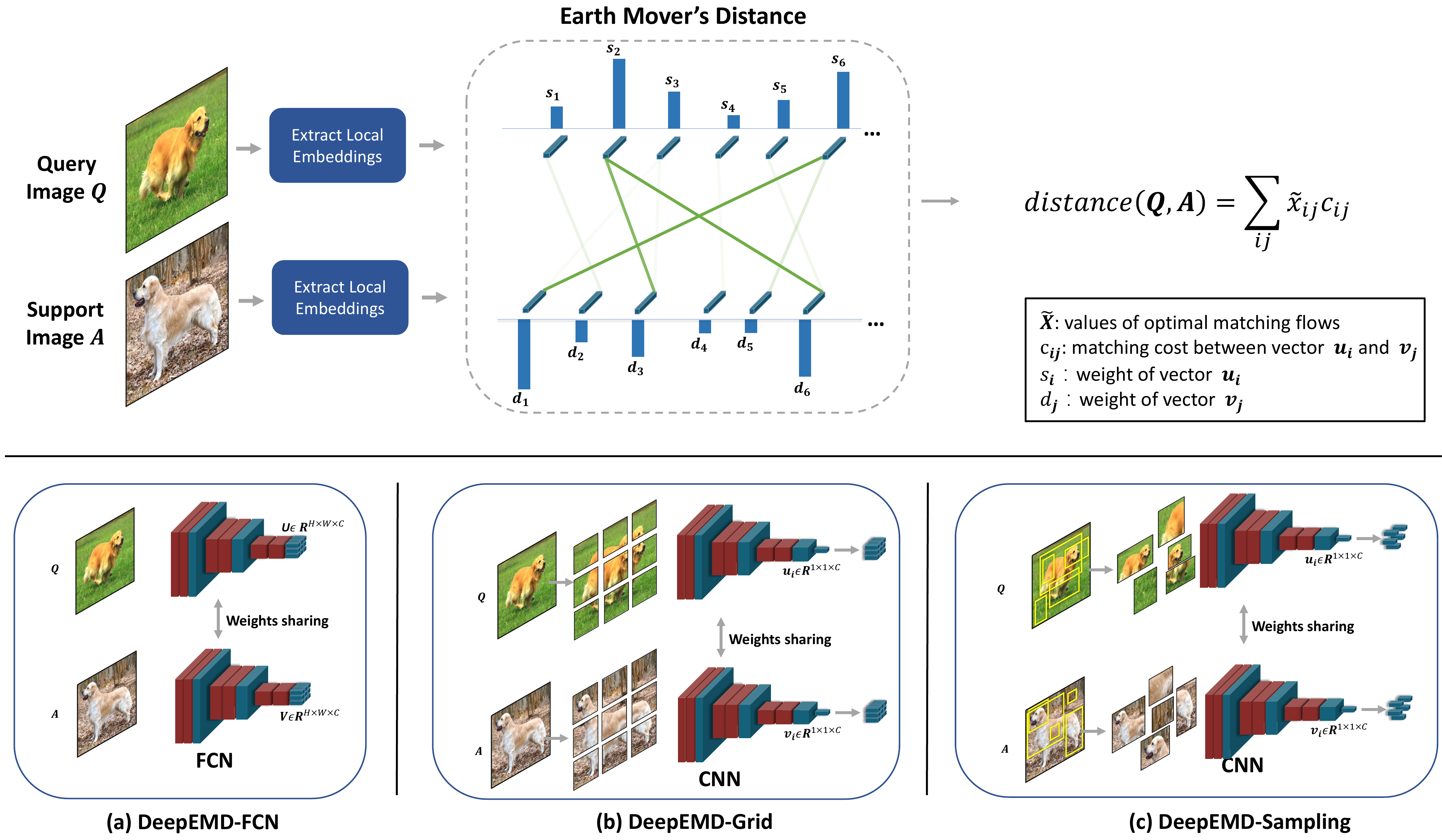}
	\caption{
	    \textbf{Our framework for 1-shot image classification}. 
	    Given a           
	    pair of images,  we first extract their local embeddings,           
	    which are two sets of feature vectors. Then we use the            
	    Earth Mover's Distance to generate the optimal matching flows between two sets, which have the minimum overall matching cost.
	    Finally, based on the optimal matching flows and matching costs, we can compute the distance between two images, which is used for classification. We explore three methods to extract local embeddings:
	    \textbf{(a)} fully convolutional networks, \textbf{(b)} cropping image patches based on grids;
	    and \textbf{(c)} random sampling of image patches. 
	    The details of the three methods are provided in Section~\ref{sec:emd}.
    }
	\label{fig:whole}
\end{figure*}

%% file: 4_0_preliminary.tex
\section{Preliminary}
Before presenting our algorithm in detail, we first introduce some preliminary concepts in the few-shot classification literature.
The general meta-learning algorithm aims to learn transferable knowledge across tasks, where knowledge learned on training tasks can be used to 
solve 
novel tasks with only a small amount of training data. 
In the few-shot classification scenario, a task $\mathcal{T}_i$ is to undertake classification over a set of sampled classes, characterized by scarce training images. Specifically, an $N$-way $K$-shot task denotes classification over $N$ classes with $K$ training samples in each class. To acquire generalization ability across tasks, the training and testing of the model are often aligned with the episodic paradigm~\cite{matchnet} where batched tasks are sampled for training or evaluation. 
For 
each sampled task, the training set $\mathcal{S}=\{(x^s_1,y^s_1),...,(x^s_{NK},y^s_{NK})\}$ is called the \emph{support set} and the testing set $\mathcal{Q}=\{(x^q_1,y^q_1),...,(x^q_{NK_Q},y^q_{NK_Q})\}$ is called the \emph{query set}, where
 $x_i$ is an image, $y_i$ is its corresponding label, $y_i \in \{1,...,N\}$, and $K_Q$ is the number of testing images per class.
At training time, the ground-truth label of the query sets provides learning supervision, and at inference time, we repeatedly sample tasks for evaluation and record their mean accuracy.

%% file: 4_method.tex
\section{Our Method} 
In this section, we first present a brief
review of the Earth Mover's Distance and describe how we formulate the one-shot classification as an optimal matching problem that can be trained end-to-end. Then, we describe our cross-reference mechanism to generate the weight of each node, which is an important parameter in the EMD formulation. Finally, we demonstrate how to use the EMD to 
%handle
tackle 
$k$-shot learning with our proposed structured fully connected layer. The overview of our framework for one-shot classification is shown in Fig.~\ref{fig:whole}.

\subsection{Revisiting the Earth Mover's Distance }

The Earth Mover's Distance is a distance measure between two sets of weighted objects or distributions, which is built upon the basic distance between individual objects and the weight of each element.
It has the form of the well-studied transportation problem from Linear Programming. 
% Specially,
Specifically, 
suppose that a set of sources or suppliers $\mathcal{S} = \{s_i \; |\; i = 1,2,...m\}$ are required to transport goods to a set of destinations or demanders $\mathcal{D} = \{d_j\;|\;j = 1,2,...k\}$, where $s_i$ denotes the supply units of
supplier $i$ and $d_j$ represents the demand of $j$-th demander. The cost per unit transported from 
supplier $i$ to demander $j$ is denoted by $c_{ij}$ , and the number of units transported is denoted by $x_{ij}$. The goal of the transportation problem is then to find a least-expensive flow of goods $\tilde{\mathcal{X}} = \{\tilde{x}_{ij} \; | \; i = 1,... m, j = 1,... k\}$ from the suppliers to the demanders:
\begin{equation}
\begin{aligned}
& \underset{x_{ij}}{\text{minimize}}
& & \sum\nolimits_{i=1}^{m}\sum\nolimits_{j=1}^{k}c_{ij}x_{ij} \\
& \text{subject to}
& & x_{ij}\geqslant 0, i=1,...,m,j=1,...,k\\
&&&\sum\nolimits_{j=1}^{k}x_{ij} = s_i,  ~ ~ i=1,...m \\
&&&\sum\nolimits_{i=1}^{m}x_{ij} = d_j, ~ ~  j=1,...k \\
\end{aligned}
\label{emd_ori}
\end{equation}
The roles of suppliers and demanders can be switched without affecting the total transportation cost. Here $s_i$ and $d_j$ are also called the weights of the nodes, which controls the total matching flows generated by each node. EMD seeks an optimal matching $\tilde{\mathcal{X}}$ between suppliers and demanders such that the overall matching cost can be minimized. The global optimal matching flows $\tilde{\mathcal{X}}$ can be achieved by solving a Linear Programming problem.

\subsection{EMD for Few-Shot Classification}
\label{sec:emd}
\begin{figure}[t]
	\centering
	\includegraphics[width=1\linewidth]{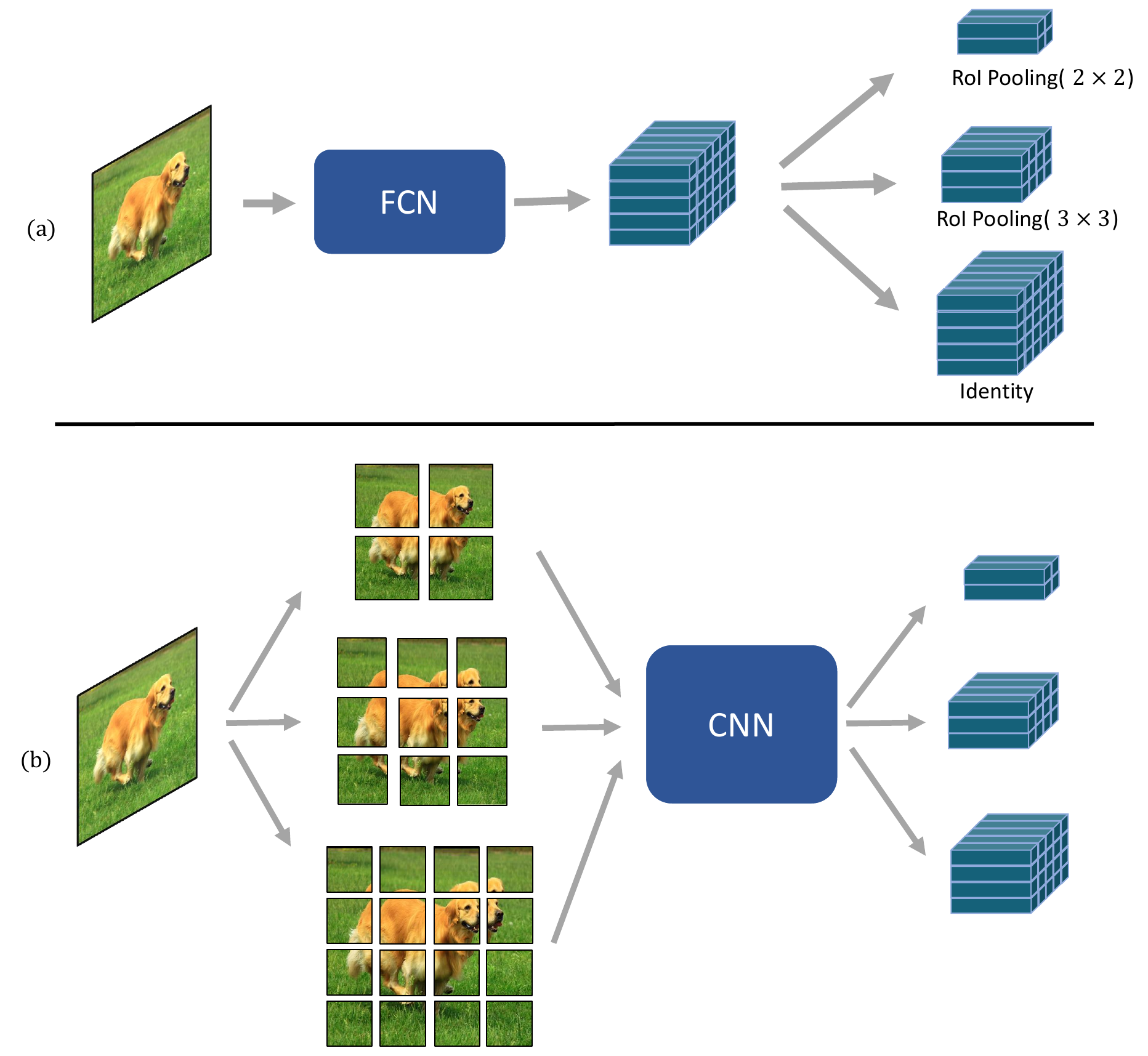}
	\caption{
	Pyramid structures applied on DeepEMD-FCN and DeepEMD-Grid to extract local embeddings. The feature pyramid structure (a) adopts RoI poolings with different output sizes on the feature maps to generate local embeddings at multiple scales while the image pyramid structure (b) crops the input image into patches according to different grid sizes, and all patches are sent to the CNN to generate local embeddings.}
	\label{fig:pyramid}
\end{figure}

In the few-shot classification task, metric-based methods aim to find a good distance metric and data representations to compute the distance between images, which are used to compare images for classification.
Different from the previous methods~\cite{proto,matchnet} that perform distance computation between the image-level embeddings, our approach advocates the use of discriminative local information. 
The intuition is that as the goal of few-shot learning is to undertake the classification task on novel categories, directly generating a category-level embedding that corresponds to a new class is difficult. On the other hand, we can decompose an object into a set of object parts that may have been seen in the training process. For example, \emph{wheel} can be a shared building element  across vehicle categories, and if such representation is learned during training, it can be useful to classify unseen vehicle categories. Therefore, local discriminative representations are likely to provide more transferable information across categories. 
In our framework, we decompose images into a set of local representations, and by assigning appropriate weights to local embeddings in two images, we can use the optimal matching cost between them to represent their dissimilarity. 
We  explore three 
% 
% methods
%
strategies 
to generate local representations from an image, as  illustrated in Fig.~\ref{fig:whole}:\\
1) \textbf{Fully Convolutional Networks.}  We can deploy a fully convolutional network (FCN)~\cite{fcn}
to generate the dense representation $\mathbf{U} \in \mathbb{R}^{H \times W \times C}$ of an image, where $H$ and $W$ denote the spatial size of the feature map and $C$ is the feature dimension.  Each image representation contains a collection of local feature vectors $\{\mathbf{u}_1,\mathbf{u}_2,...\mathbf{u}_{HW}\}$, and each vector $\mathbf{u}_i$ can be seen as a node in the set.
Thus, the dissimilarity of two images can be represented as the optimal matching cost between two sets of vectors.  \\
2) \textbf{Dividing the input image into grids.} We crop the image evenly into an $H \times W$ grid  before feeding it to the CNN, and each image patch in the grid cell is encoded by the CNN individually and generates a feature vector. The feature vectors generated by all the patches constitute the embedding set of an image.\\
3) \textbf{Random sampling 
of 
image patches.} Instead of generating the image patches by grids, here we randomly sample $M$ patches in the images with different sizes and aspect ratios. The randomly sampled patches are then re-scaled to the same input size and are encoded by CNNs. The embeddings of these sampled patches make up the embedding set of an image.\\
We denote our networks adopting the three strategies above by \textit{DeepEMD-FCN}, \textit{DeepEMD-Grid}, and \textit{DeepEMD-Sampling}, respectively. As the size of the grid in  DeepEMD-Grid and the number of patches in DeepEMD-Sampling are hyper-parameters in our network, we conduct various experiments to investigate the influence of these parameters in Section~\ref{sec:ablation}.
We also investigate pyramid structures on the image level and the feature level to capture local representations at multiple scales, illustrated in Fig.~\ref{fig:pyramid}. Specifically, we add a feature pyramid structure to DeepEMD-FCN and an image pyramid structure to DeepEMD-Grid. The feature pyramid applies the RoI pooling to the feature maps generated by the FCN, and the resulting feature vectors together with the raw feature vectors constitute the embedding set. For the image pyramid, we simply crop the patches according to different grid sizes and send all patches to the CNN to generate the embedding set.

After acquiring the embedding sets of two images, we can follow the original EMD formulation in Equation~\eqref{emd_ori} to compute the distance.
Concretely, assuming there are $H \times W$ vectors in each set, the cost per unit is obtained by computing the pairwise distance between embedding nodes $\mathbf{u}_i$, $\mathbf{v}_j$ from two image features:
\begin{equation}
c_{ij}=1-\frac{{\mathbf{u}_i}^T  \mathbf{v}_j}{\lVert \mathbf{u}_i\rVert \lVert \mathbf{v}_j\rVert},
\label{cost}
\end{equation} 
where nodes with similar representations tend to generate small matching costs between each other. As to the generation of weights $s_i$ and $d_j$, we leave the detailed elaborations in Section~\ref{sec_weight_generation}.
Once acquiring the optimal matching flows $\tilde{\mathcal{X}}$, we can compute the similarity score $s$ between image representations with:
\begin{equation}
    s(\mathbf{U},\mathbf{V})=\sum_{i=1}^{HW}\sum_{j=1}^{HW}(1-c_{ij})\tilde{x}_{ij}.
\end{equation}

\subsection{End-to-End Training}

In order to embed the optimal matching problem into a neural network for end-to-end training, it is important to make the solution of the optimal matching $\tilde{\mathcal{X}}$ differentiable with respect to the problem parameter $\theta$.
As is indicated by \cite{barratt2018differentiability}, we can apply 
the 
implicit function theorem \cite{barratt2018differentiability,dontchev2009implicit,krantz2012implicit} on the optimality (KKT) conditions to obtain the Jacobian. For the sake of completeness, we transform the optimization in  Equation~\eqref{emd_ori} to a compact matrix form: %
\begin{equation}
\begin{aligned}
& \underset{x}{\text{minimize}}
& & c(\theta)^Tx \\
& \text{subject to}
& &  G(\theta)x \leqslant h(\theta),\\
&&& A(\theta)x=b(\theta).\\
\end{aligned}
\label{lp}
\end{equation}
Here $x \in \mathbb{R}^n$ is our optimization variable, with $n = m \times k$ representing the total number of matching flows in $\mathcal{X}$. $\theta$ is the problem parameter that relates to the earlier layers in a differentiable way. $Ax = b$ represents the equality constraints and $Gx \leqslant h$ denotes the inequality constraint in Equation~\eqref{emd_ori}.
\input{sub_content/4_sparse_matrix}
Accordingly,
the Lagrangian of the LP problem in Equation~\eqref{lp} is given by:
\begin{equation}
\begin{aligned}
L(\theta,x,\nu ,\lambda) = c^Tx + \lambda^T(Gx-h)+ \nu^T(Ax-b),
\end{aligned}
\end{equation}
where $\nu$ denotes the dual variables on the equality constraints
and $\lambda \geqslant 0$ denotes the dual variables on the inequality constraints. 

Following the KKT conditions with notational convenience, we can obtain the optimum $(\tilde{x},\tilde{\nu},\tilde{\lambda})$ of the objective function by solving $g(\theta,\tilde{x},\tilde{\nu},\tilde{\lambda}) = 0$ with primal-dual interior point methods, where 
\begin{equation}
g(\theta, x,\nu,\lambda) = 
\begin{bmatrix}
\nabla_\theta L(\theta,x,\nu,\lambda)
\\ \textbf{diag}(\lambda)(G(\theta)x-h(\theta))
\\ A(\theta)x-b(\theta)
\end{bmatrix}.
\end{equation}
Then, the following theorem holds to help us derive the gradients of the LP parameters.\\
\textbf{Theorem 1} (From Barratt \cite{barratt2018differentiability}) Suppose $g(\theta,\tilde{\lambda},\tilde{\nu},\tilde{x}) = 0$. Then, when all derivatives exist, the partial Jacobian of $\tilde{x}$ with respect to $\theta$ at the optimal solution $(\tilde{\lambda},\tilde{\nu},\tilde{x})$, namely $J_\theta \tilde{x}$, can be obtained by satisfying:%
\begin{equation}%
\begin{aligned}
J_\theta \tilde{x} =& -J_x g(\theta,\tilde{\lambda},\tilde{\nu},\tilde{x})^{-1}J_\theta g(\theta,\tilde{x},\tilde{\nu},\tilde{\lambda}).%
\end{aligned}
\end{equation}
Here the formula for the Jacobian of the solution mapping is obtained by applying the implicit function theorem to the KKT conditions.
For instance, the (partial) Jacobian with respect to $\theta$ can be defined as
\begin{equation}
J_\theta g(\theta,\tilde{\lambda},\tilde{\nu},\tilde{x}) = 
\begin{bmatrix}
	J_\theta \nabla_x L(\theta,\tilde{x},\tilde{\nu},\tilde{\lambda})
	\\ \textbf{diag}(\tilde{\lambda})J_\theta (G(\theta)x-h(\theta))
	\\ J_\theta (A(\theta)\tilde{x}-b(\theta))
\end{bmatrix}.
\end{equation}
Therefore, once getting the optimal solution $\tilde{x}$ for the LP problem, we can obtain a closed-form expression for the gradient of $\tilde{x}$ with respect to the input LP parameters $\theta$. This helps us achieve an efficient backpropagation through the entire optimization process without perturbation of the initialization and optimization trajectory.

\begin{figure}[t]
	\centering
	\includegraphics[width=1\linewidth]{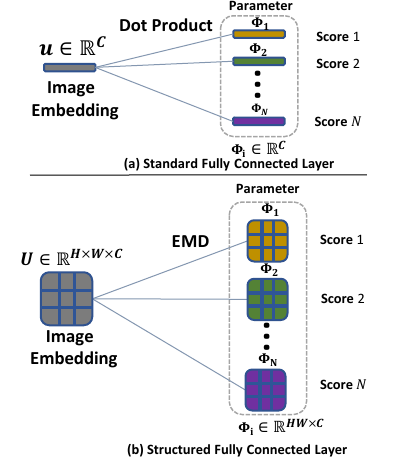}
	\caption{
Comparison of the standard fully connected layer \textbf{(a)} and our proposed structured fully connected layer (SFC) \textbf{(b)}. The SFC learns a group of vectors as the prototype for each class such that we can use the EMD to generate category scores. 
	}
	\label{fig:SFC}
\end{figure}

\subsection{Weight Generation}\label{sec_weight_generation}
As can be observed in the EMD formulation, an important problem parameter is the weight of each node, \eg, $s_i$, which controls the total matching flows $\sum_{j=1}^{k}x_{ij} $ from it. Intuitively, the node with a larger weight plays a more important role in the comparison of two sets, while a node with a very small weight can hardly influence the overall distance no matter which nodes it connects with. 
In the pioneering work that adopts EMD for color-based image retrieval~\cite{emd}, they use the histogram as the elementary feature and perform feature clustering over all pixels to generate the nodes. The weight of each node is set as the size of the corresponding cluster. It makes sense because, for color-based image retrieval, large weights should be assigned to the dominant colors with more pixels, such that the retrieved images can be visually close to the query images.
However, for few-shot image classification tasks where features for classification often contain high-level semantic information, the number of pixels does not necessarily reflect the importance. It is common to find image data with greater background regions than the target objects in classification datasets, \eg, ImageNet. Therefore, large weights should be given to the foreground object region in a matching algorithm.
However, it may be difficult to define what is the foreground region, particularly when there exist multiple object categories in a single image. An object can  be both foreground and background in different cases.
Instead of determining the weights by inspecting individual images alone,  we argue that for the few-shot classification task, the co-occurrent regions in two images are more likely to be the foreground and the weights of node features should be generated by comparing the nodes on both sides.
To achieve this goal, we propose a cross-reference mechanism that uses dot product between a node feature and the average node feature in the other structure to  generate a relevance score as the weight value:
\begin{equation}
s_i= \max \Bigl\{ 
\mathbf{u}_i^T\cdot\frac{\sum_{j=1}^{HW}{\mathbf{v}_j}}{HW},0
\Bigr\},
\end{equation}
where $\mathbf{u}_i$ and $\mathbf{v}_j$ denote the vectors from two feature maps, and function $max(\cdot)$ ensures the weights are always non-negative.  Finally, we normalize all the weights in the structure to make both sides have the same total weights for matching:
\begin{equation}
\label{eq:norm}
\hat{s}_i=s_i\frac{HW}{\sum_{j=1}^{HW}{s_j}}. %
\end{equation}
For simplicity, here we take $s_i$ as an example and $d_i$ can be obtained in the same manner.
The cross-reference mechanism aims to give less weight to the high-variance background regions and more to the co-occurrent object regions in two images. This can also put less weight on the object parts that do not co-occur in two images and thus allows partial matching to some extent. As a result, the proposed  distance metric is  based only on confident regions and confident features that have high responses.

\subsection{Structured Fully Connected Layer}
Thus far
we have discussed using the Earth Mover's Distance as the metric to generate the distance value between paired images, \ie, the one-shot case. A question is then raised---how do we tackle the $k$-shot setting where multiple support images are available?
Before presenting our design in detail, let us 
%have
% a review of
revisit how the standard fully connected layer classifies an image embedding extracted by CNNs. 
An   FC layer, parameterized by $[{\bf{\Phi}}_1,..., {\bf{\Phi}}_N]\in \mathbb{R}^{C\times N}$ contains a set of learnable vectors ${\bf{\Phi}}_i\in \mathbb{R}^{C}$ corresponding to each category. Given an image embedding $\mathbf{u}\in \mathbb{R}^{C}$ generated by the convolutional layer,
the FC layer generates the score of class $i$ by computing the dot product between the image vector $\mathbf{u}$  and the parameter vector ${\bf{\Phi}}_i$, and this process is  applied to all the categories in parallel by matrix multiplication. There are also some previous works replacing the dot product operation in the FC layer with the \emph{cosine} function for computing the category scores~\cite{closer,weightnorm}.
The learning of the FC layer can be seen as finding a prototype vector for each class such that we can use distance metrics to classify an image. An illustration of the standard FC layer is shown in Fig.~\ref{fig:SFC} (a).

With the same formulation, we can learn a structured fully connected layer that adopts EMD as the distance function  to directly classify a structured feature representation. The learnable parameter for each class becomes a group of vectors, rather than one vector, such that we can use the  structural distance function EMD to undertake image classification. This can also be interpreted as learning a prototype feature map generated by a dummy image for each class. The comparison of the structured FC and the standard FC can be found in Fig.~\ref{fig:SFC}.
Algorithm~\ref{alg:episode} provides 
the 
pseudo-code of a testing episode at inference time.
We fix the network backbone and
use SGD to learn the parameters in the structured fully connected layer by sampling data from the support set. After several iterations of optimization, we can generate the category scores by computing the EMD between the query images and each of the prototypes in the SFC.

%\FOR{<condition>} <text> \ENDFOR

% \begin{algorithm}
% \caption{An testing episode for an N-way K-shot task. N is the number of classes and K is the number of training samples in each class; $x_i$ is an image, and $y_i$ is its corresponding label where $y_i \in \{1,...,N\}$; $N_{iter}$ is the number of optimization iterations and  $N_Q$ is the number of query images.}
% \begin{algorithmic}[1]
% \REQUIRE a trained feature extractor $\Theta$, the support set $\mathcal{S}=\{(x^s_i,y^s_i)\}^{NK}_{i=1}$, the query set $\mathcal{Q}=\{(x^q_i,y^q_i)\}^{N_Q}_{i=1}$
% \ENSURE Testing accuracy $Acc$ of query images
% \label{alg:episode}
% \STATE Initialize SFC layer: $\Phi=\Phi'$
% \FOR{i \textbf{from} 1 \textbf{to} $N_{iter}$} 
% \STATE Sample a mini-batch  $\mathcal{T}_i$ from S
% \STATE make predictions for $\mathcal{T}_i$with our model $[\Theta, \Phi]$
% \STATE calculate cross entroty loss $\mathcal{L}_{\mathcal{T}}$
% \STATE optimize $\Phi$ with SGD
% \ENDFOR
% \STATE make predictions for all query set $\mathcal{Q}$
% \STATE calculate accuracy $Acc$

% \end{algorithmic}
% \end{algorithm}

\begin{algorithm}[t]
\caption{A testing episode for an $N$-way $K$-shot task. $N_{iter}$ is the number of optimization iterations.}
\label{alg:episode}
\SetAlgoLined
\SetKwInput{KwData}{Input}
\SetKwInput{KwResult}{Output}
 \KwData{a trained feature extractor $\Theta$, the support set $\mathcal{S}$, and the query set $\mathcal{Q}$}
 \KwResult{Testing accuracy $Acc$ of query set.}
 Initialize SFC layer: $\Phi=\Phi'$\;
 \For{i \textbf{from} 1 \textbf{to} $N_{iter}$}{
 Sample a mini-batch  $\mathcal{B}_i$ from S\;
 make predictions for $\mathcal{B}_i$ with our model $[\Theta, \Phi]$\;
 calculate cross entropy  loss $\mathcal{L}_{\mathcal{T}}$\;
 optimize $\Phi$ with SGD\;
  }
 make predictions for images in the query set $\mathcal{Q}$\ with the model $[\Theta, \Phi]$\;
calculate accuracy $Acc$\;
return  $Acc$.
\end{algorithm}
% \vspace{-6mm}

%% file: sub_content/4_sparse_matrix.tex
Specifically, to construct the compact matrix form of the original optimization, we can build up the sparse matrix below for the equality constraints: %$c \in \mathbb{R}^n$, $G \in \mathbb{R}^{n \times n}$, $h \in \mathbb{R}^n$, $A \in \mathbb{R}^{(m+k) \times n}$, $b \in  \mathbb{R}^{m+k}$. 
\begin{equation}
\begin{bmatrix}
1 & \cdots  &  1&  &  &  &  &  & \\ 
&  &  &  1& \cdots  & 1 &  &  & \\ 
&  &  &  &  &  & 1 &  \cdots & 1\\ 
1&  &  &  1&  &  &  1&  & \\ 
%&  1&  &  &1  &  &  & 1 & \\ 
& \ddots &   &  &\ddots  &  &  & \ddots  & \\ 
&  & 1 &  &  &  1& &  & 1\\ 
\end{bmatrix} 
\begin{bmatrix}
	x_{11}\\ 
	x_{12}\\ 
	\vdots \\ 
	x_{1k}\\ 
	x_{21}\\ 
	x_{22}\\ 
	\vdots \\ 
	x_{2k}\\ 
	\vdots \\ 
	x_{m1}\\ 
	x_{m2}\\ 
	x_{mk}\\ 
	
\end{bmatrix}
=
\begin{bmatrix}
	s_1\\ 
	s_2\\ 
	\vdots \\ 
	s_m\\ 
	d_1\\ 
	d_2\\ 
	\vdots \\ 
	d_k
\end{bmatrix},
\end{equation}
and the inequality constraint can be written as:
\begin{equation}
\begin{bmatrix}
-1 &  &  &  & \\ 
&  -1&  &  & \\ 
&  &  \ddots &  & \\ 
&  &  &  -1& \\ 
&  &  &  & -1
\end{bmatrix}
\begin{bmatrix}
x_{11}\\ 
x_{12}\\ 
\vdots \\ 
x_{1k}\\ 
x_{21}\\ 
x_{22}\\ 
\vdots \\ 
x_{2k}\\ 
\vdots \\ 
x_{m1}\\ 
x_{m2}\\ 
x_{mk}\\ 
\end{bmatrix}
\leqslant
\begin{bmatrix}
0\\ 
0\\ 
0\\ 
\vdots \\ 
0\\ 
0\\ 

\end{bmatrix}.
\end{equation}

%% file: 5_experiment.tex
\section{Experiments}
To evaluate the performance of our proposed algorithm for few-shot classification, we conduct extensive experiments on multiple datasets. In this section, we first present dataset information and some important implementation details in our network design. Then we conduct various ablative experiments to validate the effectiveness of each component in our network and compare our model with the state-of-the-art methods on popular benchmark datasets. Finally, we validate the effectiveness of our model on the image retrieval task.

\subsection{Implementation Details}
\textbf{Network.} For a fair comparison with previous works, we employ a 12-layer ResNet (ResNet12) as our model backbone, which is widely used in the few-shot classification literature. For DeepEMD-Grid and DeepEMD-Sampling, we remove the fully connected layer in ResNet, such that the network generates a vector for each input image patch. For DeepEMD-FCN, we further remove the global average pooling layer, such that the network is transformed into a  fully convolutional network. Specifically, given an image of size $84\times84$, the model generates a feature map of size $5\times5\times512$, \ie, 25 feature vectors. For DeepEMD-Grid, we slightly enlarge the region of the local patches in the grid by a factor of 2 to incorporate context information, which is found helpful to generate local representations. 

\textbf{Training.} At training time, we use the GPU accelerated convex optimization solver QPTH~\cite{amos2017optnet} to solve the Linear Programming problem in our network and compute gradients for back-propagation. 
As is commonly implemented in the state-of-the-art literature~\cite{feat}, we adopt a feature pre-training step followed by  episodic meta-training~\cite{matchnet} to learn our network.
At the pre-training stage, we train a standard classification model with all training classes and we use DeepEMD-FCN for validation with the validation set in the training process. After pre-training, the model with the highest validation accuracy is further optimized by episodic training for 5,000 episodes.  
In each training episode, we randomly sample a 5-way 1-shot task with 16 query images, which is aligned with the testing episodes.
For the \emph{k}-shot classification task, we re-use the trained 1-shot model as the network backbone to extract features and fix it during training and testing.
We initialize the parameters in the structured FC layer with the average local representations of all support data in each class and sample a mini-batch of 5 images from the support set to finetune the structured FC layer for 100 iterations.

\input{sub_content/5_analysis_experiment}

\subsection{Dataset Description}
We conduct few-shot classification experiments on five popular benchmark datasets, 
namely, 
\emph{mini}ImageNet \cite{matchnet}, \emph{tiered}ImageNet  \cite{tieredimagenet}, Fewshot-CIFAR100 (FC100) \cite{TADAM}, Cal\-tech-UCSD Birds-200-2011 (CUB) \cite{cub}, and CIFAR-Few\-Shot (CIFAR-FS) \cite{bertinetto2018meta}.

\textbf{\emph{mini}ImageNet.} \emph{mini}ImageNet was first proposed in~\cite{matchnet} and becomes the most popular benchmark in the few-shot classification literature. It contains 100 classes with 600 images in each class, which are built upon the ImageNet dataset~\cite{imagenet}. The 100 classes are divided into 64, 16, and 20 for meta-training, meta-validation, and meta-testing, respectively.

\textbf{\emph{tiered}ImageNet.} \emph{tiered}ImageNet is also a subset of ImageNet, which includes 608 classes from 34 super-classes. Compared with  \emph{mini}ImageNet, the splits of meta-training(20), meta-validation(6), and meta-testing(8) are set according to the super-classes to enlarge the domain difference between  training and testing phases. The dataset also includes more images for training and evaluation (779,165 images in total).

\textbf{Fewshot-CIFAR100.} FC100 is a few-shot classification dataset built on CIFAR100~\cite{CIFAR100}. We follow the split division proposed in~\cite{TADAM}, where 36 super-classes were divided into 12 (including 60 classes), 4 (including 20 classes), and 4 (including 20 classes), for meta-training, meta-validation, and meta-testing, respectively, and each class contains 100 images.

\textbf{CIFAR-FewShot.} CIFAR-FS~\cite{bertinetto2018meta} is also a few-shot classification dataset built on CIFAR100~\cite{CIFAR100}. It contains 64, 15, and 20 classes for training, validation, and testing, respectively.

\textbf{Caltech-UCSD Birds-200-2011.} CUB was originally proposed for fine-grained bird classification, which contains 11,788 images from 200 classes. We follow the splits in~\cite{feat} where 200 classes are divided into 100, 50, and 50 for meta-training, meta-validation, and meta-testing, respectively.

\subsection{Analysis}
\label{sec:ablation}
In this %part,
section, 
we implement various experiments to evaluate the effectiveness of our algorithm. We also explore multiple design variants of our network and compare them with baseline solutions. All the experiments are conducted on the
\emph{mini}ImageNet dataset. 

\textbf{Comparison with methods based on image-level representations.} In the beginning, we first compare our method with a set of methods that utilize image-level vector representations on the 1-shot task. These methods maintain the global average pooling operation in ResNet to generate vector representations for images and use various distance metrics for classification. We select the representative  methods in the literature for comparison: 1) Prototypical Network~\cite{proto} with Euclidean distance. 2) Matching Network~\cite{matchnet} with \emph{cosine} distance. 3) Finetuning a FC classifier. In~\cite{closer}, Chen \etal propose to fix the pretrained feature extractor and finetune the FC layer with the support images.
For fair comparisons, we adopt the same backbones and training schemes for all these baseline methods. 
We use FCN to extract local features in this experiment such that the only difference in the backbone is the lack of global average pooling operation in our model.
The experiment result is shown in Table~\ref{table:1shotcomparison}. As we can see, our algorithm significantly outperforms baseline methods that rely on image-level vector representations under both 1-shot 5-way and 1-shot 10-way settings, which validates the effectiveness of the optimal matching based method that relies on local features.

\textbf{Comparison with methods based on local representations.} There are also a few methods in the literature focusing on local representations to solve few-shot classification. They all remove the global average pooling in the CNN to obtain dense representations of images. In~\cite{Revisiting}, Li \etal use the top $k$ nearest vectors (\textbf{KNN}) between two feature maps to represent the image-level distance. Lifchitz \etal~\cite{dense} propose to make predictions with each local representation and average their output probabilities. We replace our EMD head with their methods for comparison.  The result is shown in Table~\ref{table:1shotcomparison}. Our optimal matching based algorithm outperforms all other model variants. Compared with other methods based on local features, the advantage of our method is that,  although the basic ground distance in the EMD is based on local features, our algorithm compares the two structures in a global way. Predictions solely based on nearest local features in two images may not extract sufficient information to differentiate between images. For example, \emph{eyes} can be the nearest feature between  animal images, but such a feature can hardly be used to differentiate between animal species.

\input{sub_content/table_pyramid}

\input{sub_content/table_ss}

\input{sub_content/table_cross}

\input{sub_content/table_time}

\textbf{Weights in the EMD.} We next investigate the influence of weights in the EMD. We use the FCN to extract local embeddings for all the methods in this experiment.
The first baseline is to set equal weight values for all local embeddings, which is denoted by \textbf{Equal}.
The vanilla EMD~\cite{emd} for image retrieval uses the pixel color as the feature and clusters pixels to generate nodes. The weight of the node is set with the portion of pixels in this cluster.
We examine two clustering algorithms as baselines to generate weights: K-means~\cite{kmeans} and Mean-shift~\cite{meanshift} which are implemented by the Scikit-learn~\cite{scikit-learn} library.
For both clustering algorithms, we use the cluster mean as the local embedding and the cluster size, \ie, the number of feature vectors in each cluster, as the weight, for computing EMD.
It is important to note  that there are two special cases that deviate from the goal of using clustering algorithms to set the weights.
The first case is that each individual local embedding is a cluster, which means no clustering is applied. This corresponds to K-means with K = 25, and Mean-shift with bandwidth (\textbf{BD})  $ \approx 0$. As a result, there are totally 25 clusters of an image with equal weights assigned to each cluster. This amounts to our first baseline that sets equal weights in EMD.
The second case is that all local embeddings of an image are clustered into one cluster. This corresponds to K-means with K = 1, and Mean-shift with BD $ \approx \infty$.  In this case, an image is represented by a single vector, which is the cluster mean, and the EMD is no longer a structured distance function and becomes the cosine distance between two global vector embeddings, which amounts to the baseline in Table~\ref{table:1shotcomparison}.
Therefore, for K-means, we choose the number of clusters from $\{25, 10, 5, 2, 1\}$. 
For Mean-shift, we manually select multiple values to set the hyper-parameter, bandwidth, from $\{0, 2.5, 5, 7.5, 10, 12.5, 15, \infty\}$. 
We also use the automatically estimated bandwidth value for each image provided in the Scikit-learn~\cite{scikit-learn} library. To better observe the influence of the bandwidth values, we additionally list the average number of clusters generated by Mean-shift in each image during testing.
As the clustering process of the  aforementioned algorithms is non-differentiable, for a fair comparison, \emph{we fix the parameters in the backbone after pre-training to evaluate all methods}.
To further test whether our performance advantage is solely brought by the cross-reference mechanism, 
we also compare our network with a model variant that is solely based on the cross-reference mechanism without EMD. Concretely, we compute the \emph{cosine} distance between all vector pairs, denoted by \textbf{Dense Connections}, and compute a weighted sum of these distances with the node weights generated by the cross-reference mechanism.
This baseline draws connections with CrossTransformers~\cite{crosstrans} and Cross Attention Networks~\cite{CAN}, where spatial attentions based on local correspondence between samples are  used to refine the data embeddings for better classification. Here we use the  cross-reference mechanism  to  weight the distances between local embeddings, which plays a similar role with attentions. 

\begin{figure}[t]
	\centering
	\includegraphics[width=1\linewidth]{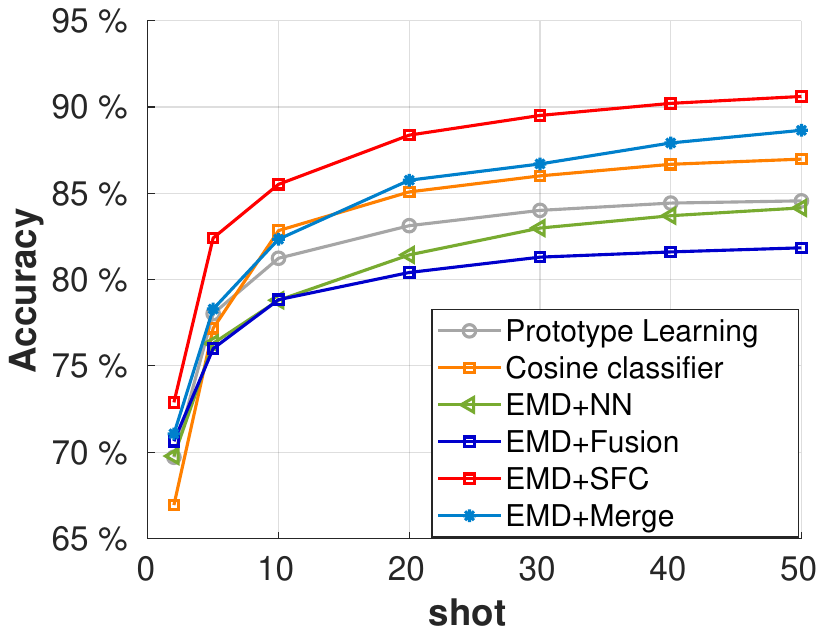}
	\caption{Experiment on 5-way $k$-shot classification. The proposed structured FC layer significantly outperforms baselines and previous $k$-shot solutions.}
	\label{fig:kshot}
%\vskip -1em
\end{figure}

As we can see from the results in Table~\ref{table:weights}, our cross-reference mechanism can bring an improvement of up to 4.2\% over the baseline with equal weights, while using the clustering-based methods to set the weights can not improve the performance, which validates our hypothesis that the number of pixels does not necessarily correspond to the importance in few-shot classification. 
The optimal results of  K-means and Mean-shift are both obtained in the second special case, where all local embeddings of an image are clustered into one cluster. This indicates that EMD with weights set by clustering algorithms is inferior to cosine distance with global representations.
We find that for clustering-based models, better results are always obtained when the clustering results are close to the two special cases described above. Using the auto-estimated bandwidth for Mean-shift results in the poorest results, although the average number of generated clusters looks normal, which indicates that a good clustering result does not necessarily lead to better classification performance.
For the model variant solely based on the cross-reference mechanism as an attention, it can only slightly improve the result of the simple average operation, while a combination of the cross-reference mechanism and the EMD can yield a significant performance improvement, which again validates the advantages of using the EMD as the metric and the effectiveness of the cross-reference mechanism.

\begin{figure*}[]
	\centering
	\includegraphics[width=1\linewidth]{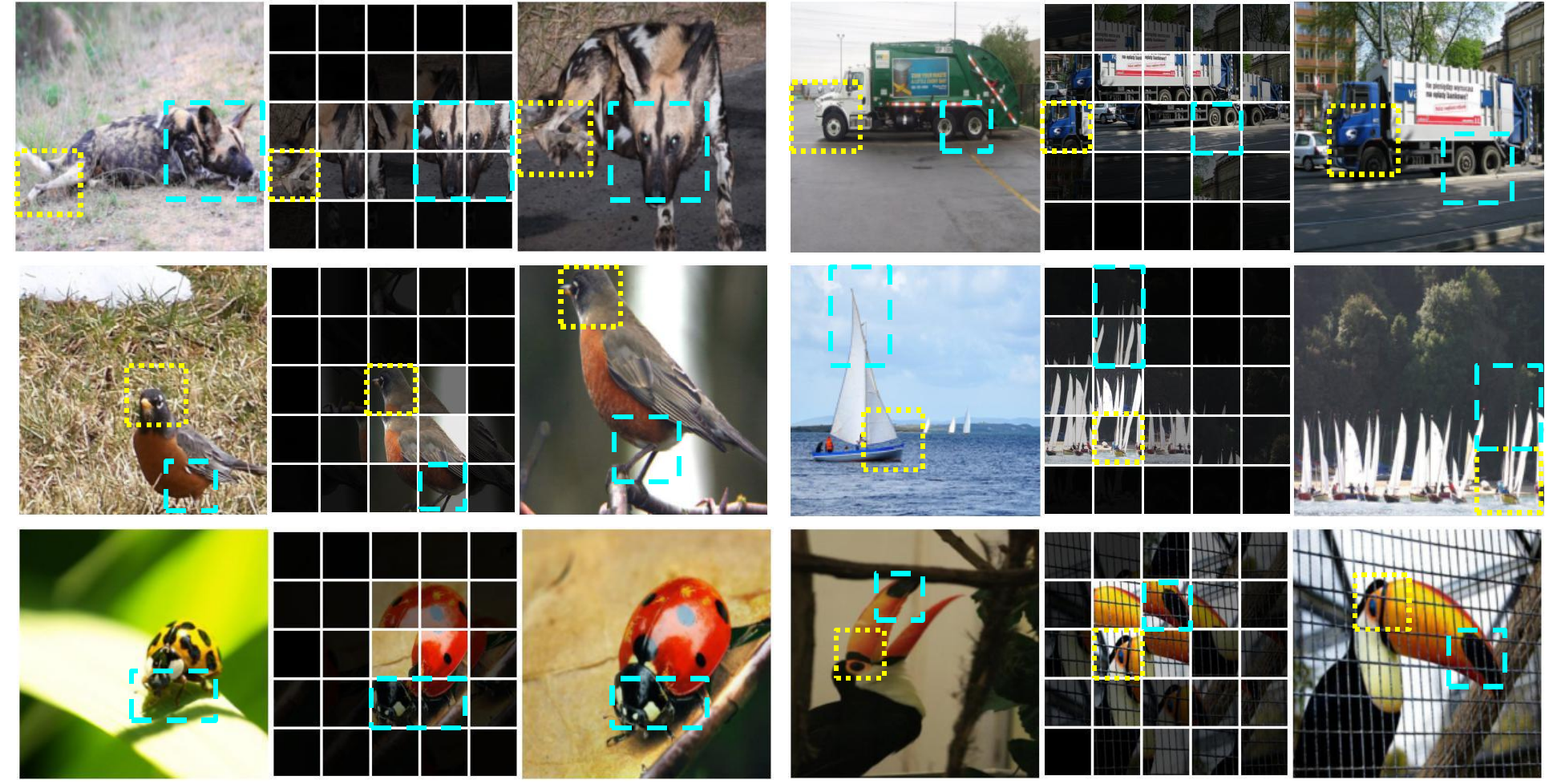}
	\caption{ Visual reconstruction experiment by DeepEMD-FCN. Given two images (left and right), we plot in the middle the best matched patch (the one with the highest flow value) of each local region in the left image.  It can  be seen as the reconstruction of the left image using patches from the right one.
	The weight controls the brightness of the corresponding region.  Our algorithm can effectively establish semantic correspondence between local regions and assign small weights to the background regions.   }
	\label{fig:Visualization}
\end{figure*}

 \begin{figure*}[t]
	\centering
	\includegraphics[width=0.9\linewidth]{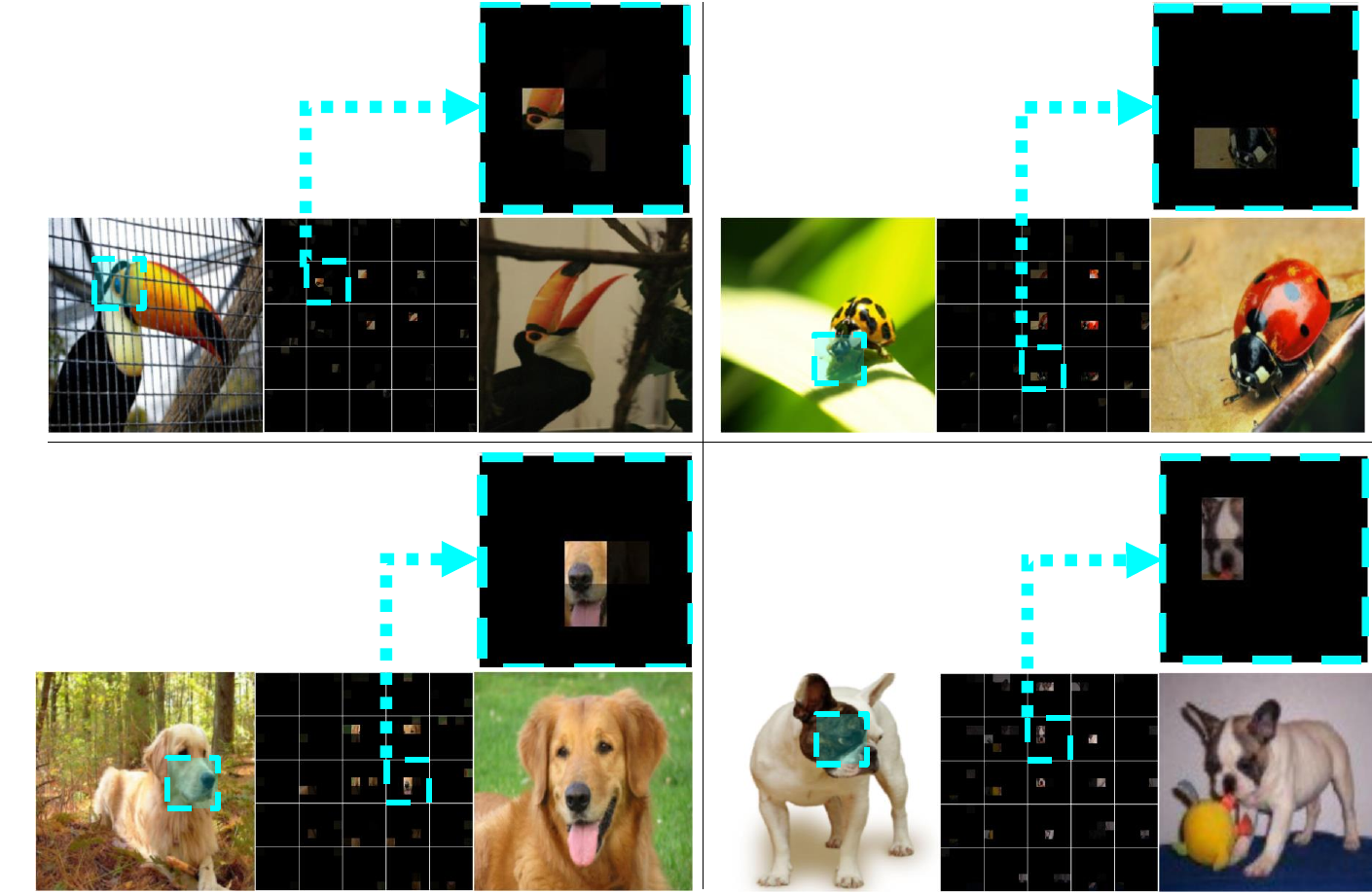}
	\caption{ 
	Visualization of full matching flows in DeepEMD-FCN. Each grid cell in the middle contains the matched patches (from the right image) of the corresponding region in the left image. The brightness is controlled by the flow values and the weights. \textit{Please zoom in for details}. }

	\label{fig:Visualization-full}
\end{figure*}

\input{6_1_soatable.tex}

\textbf{Local embedding extractor.} In Table~\ref{table:extractor}, we compare the three methods to extract local embeddings of the input image described in Section~\ref{sec:emd}.
We also investigate some key parameters in respective methods, \eg the size of grids in DeepEMD-Grid and the number of patches in DeepEMD-Sampling. From the experiment result, we have the following findings:\\
1) DeepEMD-Sampling shows distinct advantages over the plain versions of the other two methods.
For DeepEMD-Sampling, increasing the number of sampled patches can consistently boost the performance, even if more low-quality patches, \eg patches from the background region, are likely to be sampled. 
We observe the weights of these patches and find that they are usually assigned with small weights, thanks to our proposed cross-reference mechanism. Therefore, these noisy patches contribute less to the overall distance, and increasing the number of sampled patches does not negatively influence the prediction. \\
%Sampling duplicate patches does not influence the overall distance either, as the weights of all patches are normalized in Equation~\ref{eq:norm} and two identical nodes can be fused into one without    
2) Increasing the number of cells in DeepEMD-Grid results in small input image patches and the performance  degrades. 
A possible explanation is that small image patches decompose the object in the image into small pieces, which may lose context information and raises the difficulty in generating high-level representations.\\
3) We speculate that the advantage of DeepEMD-Sampling in the performance over other two methods is gained by multi-scale information implicitly captured by random sampling, as the embedding set contains randomly sampled local descriptors that cover different sizes of local regions. Moreover, as all sampled patches are resized to the input size, an object may appear at an appropriate size in the input image, which can have a stronger feature response in the CNN. 
% The influence of multi-scale information is widely seen in computer vision tasks, \eg object detection and image segmentation.
To validate our hypothesis, we add pyramid structures to the other two methods to incorporate multi-scale information, as illustrated in Fig.~\ref{fig:pyramid}.  Then the new embedding sets generated by the pyramid structures are used for computing the EMD. As is shown in Table~\ref{table:extractor}, the pyramid structures can effectively boost the performance of DeepEMD-FCN and DeepEMD-Grid, which indicates that multi-scale information is useful for reasoning the relations between images in the low-data regime, and pyramid structures can work at both the image level and the feature level.

\textbf{Comparison with other \emph{k}-shot methods.}
We next compare our proposed SFC layer (denoted by \textbf{EMD + SFC} ) with some baseline methods and other \emph{k}-shot solutions in the literature.
\begin{itemize}
    \item

\textbf{EMD + NN}.
As the EMD  is a paired function for two structures, the first baseline model for $k$-shot experiment is the nearest neighbor  (\textbf{NN}) method.  We compute the Earth Mover's distance between the query image and all support images and then classify the query image as the category of the nearest support sample.

\item 
\textbf{EMD + Fusion}. Instead of taking the label of the nearest sample, we can fuse the distances of all support images belonging to the same class and classify the query image to the class with the minimum overall distance. 

\item 
\textbf{EMD + Merge}. As the number of elements in the compared sets can differ, we can also merge the local embeddings of all support samples in each class into a big embedding set to compute EMD, and there will be $k \times 5 \times 5$  embedding vectors in the set. The problem in this baseline is that as the size of the set grows linearly with respect to the number of shots, the time complexity of solving the transportation problem increases cubically with the QPTH solver, which makes the inference of many-shot tasks very slow. Please refer to Section~\ref{sec:time} for the analysis of time complexity.

\item 
\textbf{Prototype Learning}. In~\cite{proto}, they average the feature embeddings of support images belonging to the same class as the category prototype and apply the nearest neighbor  method for classification.

\item 
\textbf{Finetuning a cosine classifier~\cite{closer}}. Similar to our proposed SFC layer, the network backbone is fixed and the support images are used to learn a fully connected layer as the classifier. The difference is the parameters of the prototypes and the distance metrics for classification, illustrated earlier in Fig.~\ref{fig:SFC}.

\end{itemize}
We compare these models on the $k$-shot 5-way tasks with multiple $k$ values, and the results are shown in Fig.~\ref{fig:kshot}. We can find that the performance of non-optimization based methods often gets saturated quickly and the performance increases slowly with more support samples available.
Our structured FC layer can consistently outperform baseline models, and with the number of support images increasing, our network shows even more advantages. The comparison between our method and the finetuned \emph{cosine} classifier further shows the advantages of structured prototypes and EMD as the metric in the proposed SFC layer.

\textbf{Pre-training with self-supervised auxiliary tasks.} 
As the weight generation and the computation of optimal matching flows rely on the representations encoded by the backbone, the feature pre-training plays an important role in our framework.
We next investigate the use of self-supervision  at the pre-training stage.
Self-supervised learning~\cite{moco,mangla2020charting} has been recently employed as a pretext task to learn generic representations, which benefits many downstream vision tasks, such as object detection and instance segmentation.
Following~\cite{mangla2020charting}, we add a self-supervised auxiliary learning task during backbone pre-training and observe its influence on our method.
Specifically, we randomly rotate the input images by $r$ degrees, and $r\in \{0^{\circ}, 90^{\circ}, 180^{\circ}, 270^{\circ} \}$.
The self-supervised learning task is to predict which rotation angle is applied to each image. More concretely, we add a 4-way linear classifier after the network backbone to predict the selections, as is done in~\cite{mangla2020charting}. The auxiliary cross-entropy loss is weighted by a factor of $\gamma=0.5$, which is searched based on the validation accuracy. As can be seen from the result in Table~\ref{table:ss}, the self-supervised auxiliary task at the feature pre-training stage can effectively boost the performance of our proposed method by up to 1.3\%, while applying the random rotation as data augmentation alone degrades the performance in many tasks. This indicates that that class-agnostic self-supervised learning task is beneficial to the representation learning in our framework.

\subsection{Cross-Domain Experiments}
Following the experimental setups in~\cite{closer}, we perform a cross-domain experiment where models are trained on \emph{mini}Imagenet and evaluated on the CUB dataset. Due to the large domain gap, we can better evaluate  the models' ability to tackle large domain difference between tasks.
We compare our proposed method with baseline models in the Table~\ref{table:cross}. As we can see, our algorithm \textit{ outperforms the baseline models with a large margin}. This shows  that local features can provide more transferable and discriminative information across domains. Moreover, due to our cross-reference mechanism, the optimal matching can be restricted within the co-occurrent object regions that have high feature response, such that the final distance is based on confident regions and representations, and thus has the ability to filter noise when there is a huge domain shift.

\subsection{Visualization of Matching Flows and Weights}
It is interesting to visualize the optimal matching flows and node weights in the network inference process.
We conduct two visualization experiments: 
First, since we have the correspondence information between the regions in two images, we can reconstruct  one image with the local patches from the other image. In Fig.~\ref{fig:Visualization}, we paste the best-matched patches from the right image to the corresponding position in the left image, and the weights of patches control the brightness of the  regions.  As we can see, our algorithm can effectively establish semantic correspondence between local regions, and the background regions in the left image are assigned with small weights, thus contributing less to the overall distance.
Second, as the correspondence between regions is not strictly 1-to-1, we then plot the full optimal matching flows in  Fig.~\ref{fig:Visualization-full}. As is shown, one patch can  be related to multiple regions in the other image with different weights, which is a useful property when the sizes of the same object are different in two images.

\subsection{Time Complexity}
\label{sec:time}
Compared with the methods with a closed-form distance metric, the training and inference of DeepEMD come with more computation cost, as an LP problem must be solved for each forward pass. As is discussed in~\cite{amos2017optnet}, the main computation lies in the factorization of the KKT matrix as well as back-substitution when using the interior point method to solve the LP problem, which have cubic and quadratic time complexity respectively with respect to the number of optimization variables. The depth of the backbone network has negligible influence on the computation time of the EMD layer, as only the weight of each vector and ground distance matrix are needed for computing EMD, and they can be  efficiently computed in parallel in modern tensor processing libraries, \eg PyTorch, TensorFlow, and NumPy.
Therefore, for DeepEMD-FCN, poolings are helpful in reducing computation time when a feature map with a large spatial size is generated by the backbone. 
In Table~\ref{table:time}, we compare the computation time of the EMD layer in a forward pass of a 5-way 1-shot task, given different sizes of feature maps and dimensions.
At training time, we use the QPTH library which adopts the interior point method  to solve the LP problem. As we can see, the feature dimension has little influence on the computation time, which indicates that we can replace the backbone with deeper ones, \eg, ResNet-101 (2048 channels), without significantly increasing the inference time.  As the interior point method is only necessary for computing gradients  at training time, after the model is trained, we can replace the solver with other solvers that can make faster inference, \eg, Sinkhorn~\cite{cuturi2013sinkhorn} and Simplex. At inference time, we deploy the model with the OpenCV~\cite{OpenCV} library, which adopts a modified Simplex algorithm to solve the LP problem. As is shown in Table~\ref{table:time}, the Simplex solver is much faster than interior point method. Therefore, we use QPTH to train the network and use OpenCV to deploy a trained model for inference.

\subsection{Comparison with  State-of-the-art Methods}
Finally, we compare our algorithm with the state-of-the-art methods. We report 1-shot 5-way and 5-shot 5-way performance on 5 popular benchmarks: \emph{mini}ImageNet, \emph{tiered}ImageNet, FC100, CUB and CIFAR-FS. For the 1-shot experiment, we repeatedly sample 5,000 testing episodes and record their average accuracy, and for 5-shot experiments, we sample 600 episodes.
We reproduce the methods in some earlier works~\cite{closer,matchnet,proto} with our network backbone and training strategies, and report the higher performance between our results and their reported ones. The results are shown in Table~\ref{table:soa}.
\textit{Our algorithm achieves new state-of-the-art performance on all datasets without seeking any extra data.  In particular, our results outperform the state-of-the-art performance by a significant margin on multiple tasks}, \eg,
1-shot (\textbf{3.47\%}) and 5-shot (\textbf{1.43\%}) on the \emph{mini}ImageNet dataset; 
1-shot (\textbf{2.77\%}) and 5-shot (\textbf{1.05\%}) on the \emph{tiered}ImageNet dataset.
We observe that on the FC100 and CIFAR-FS datasets, DeepEMD-FCN outperforms DeepEMD-Grid and DeepEMD-Sampling, which is different from the observations on other datasets. The possible reason is that these two datasets are built upon CIFAR100, where the size of images is $32 \times 32$ . Further cropping patches from the image results in very small input images, which makes the CNNs difficult to generate useful representations.

\subsection{Experiments on Image Retrieval}
We next validate the effectiveness of our method on the image retrieval task, which is also evaluated based on the pairwise image similarity.
Different from most existing works in the deep metric learning literature that focus on the strategy of optimization and losses, our method provides a generic metric that is complementary to many existing deep metric learning algorithms for images.
We observe the gain on the performance when we combine some high-performing baseline methods with our DeepEMD. 
To do so, we simply replace the globally pooled vector representations with local representations and replace the metric with  DeepEMD.

\textbf{Details.} Our experiment mainly follows a recent comprehensive empirical study in~\cite{musgrave2020metric}, where existing deep metric learning algorithms are fairly evaluated by three metrics based on image similarities, including, Recall@1 (\textbf{P@1}), R-Precision (\textbf{RP}), and Mean Average Precision at R (\textbf{MAP@R}).  The experiment is conducted on CUB dataset, where the first 100 classes are used for training, and the rest classes are used for evaluation.
We use ImageNet pre-trained ResNet-50 as the backbone, and a linear layer is added to project the output embedding to 128 channels.
The training and evaluation configurations are kept same for the baselines and our method for fair comparisons.
% we additionally add a pre-trained-only baseline, where we directly use the ImageNet pre-trained  ResNet-50 without further training to make predictions on the CUB dataset.
For DeepEMD-\textbf{F}CN, we use the feature pyramid of \{3, 2, 1\}; for DeepEMD-\textbf{G}rid, we use the image pyramid of \{3, 2\}; for DeepEMD-\textbf{S}ampling, we randomly sample 25 patches. Please refer to~\cite{musgrave2020metric} for more details about the evaluation metrics and baselines.

\textbf{Results.} The comparisons between baselines and our methods are shown in Table~\ref{table:metric}. As we can see, our method can effectively improve the performance of baselines under three evaluation metrics.  
% In particular, the improvement over the ImageNet pre-trained-only model  is significant, which again validate the effectiveness of our algorithm in the cross-domain case.

\input{sub_content/table_metric}

%% file: sub_content/5_analysis_experiment.tex
\begin{table}[t]
\centering

\caption{Comparison with different baseline methods  for 1-shot classification. Our model with EMD as the distance metric significantly outperforms baseline models based on image-level representations and local representations. }

\resizebox{0.47\textwidth}{!}{%
\begin{tabular}{lcccc}
\toprule[1.2pt]
\multicolumn{1}{l}{Model}   & Embedding & Metric  &  5-way &10-way\\ \hline\hline
ProtoNet~\cite{proto} & global &Euclidean &60.37  &44.34         \\ 
MatchingNet~\cite{matchnet}& global  & \emph{cosine} &63.08 &47.09               \\
FC~\cite{closer} & global &  \emph{dot} &59.41      &44.08    \\ 
FC~\cite{closer}  & global & \emph{cosine} &55.43  &40.42         \\ 
\Xhline{1pt}
KNN~\cite{Revisiting}  &local&\emph{cosine}         &62.52&47.08  \\ 
Prediction Fusion~\cite{dense} & local & \emph{cosine} &62.38   &47.04             \\
\textbf{DeepEMD-FCN (our)} &local & EMD & \textbf{65.91} & \textbf{49.66} \\  \Xhline{1.2pt}
\end{tabular}%
}

\label{table:1shotcomparison}
\end{table}

\begin{table}[t]
\centering

\caption{Different methods for setting the weights in the EMD.  We report the 1-shot performance with only the feature pre-training step. DC denotes dense connections. $K$ denotes the number of clusters in K-means; BD denotes the bandwidth in Mean-shift; $mK$ denotes the average number of clusters generated by Mean-shift.  EMD with our cross-reference (\textbf{CR}) mechanism yields the best result. The model variant that is solely based on the cross-reference mechanism as attention without EMD causes a significant performance drop.  }

\small 
\resizebox{0.47\textwidth}{!}{%
\begin{tabular}{llcc}
\toprule[1pt]
\multicolumn{1}{c}{Method}  & Operation & 5-way& 10-way \\ \hline\hline
DC&Average & 55.16  & 40.88         \\ 
DC&\textbf{CR} & 55.41  & 41.60         \\ 
\Xhline{1pt}
EMD&Equal & 56.95  & 42.89         \\ \hline
EMD&K-means ($K=25$)  & 56.95  & 42.89             \\
EMD&K-means ($K=10$)  & 56.25   & 41.85             \\
EMD&K-means ($K=5$)  & 55.92   & 41.57             \\
EMD&K-means ($K=2$)  & 56.02   & 41.75             \\
EMD&K-means ($K=1$)  & 58.65   & 44.13             \\\hline
EMD&Mean-shift (BD $ \approx 0, mK=25$) &  56.95  & 42.89           \\
EMD&Mean-shift (BD $= 2.5, mK=24.7$) & 56.93   &   42.85         \\
EMD&Mean-shift (BD $= 5, mK=22.8$) &56.66    &    42.63        \\
EMD&Mean-shift (BD $= 7.5, mK=18.1$) &  56.28  &  42.13         \\
EMD&Mean-shift (BD $= 10, mK=10.6$) &  54.99  &    40.55        \\
EMD&Mean-shift (BD $= 12.5, mK=4.3$) &   54.18 &    40.21        \\
EMD&Mean-shift (BD $= 15, mK=1.8$) & 55.48    &      41.80     \\
EMD&Mean-shift (BD $ \approx \infty, mK=1$) & 58.65   & 44.13            \\
EMD&Mean-shift (auto BD, $mK=3.4$)& 53.56 & 39.70               \\\hline

EMD& \textbf{CR} & \textbf{61.13}  & \textbf{46.92}              \\
 \Xhline{1.2pt}
\end{tabular}%
}
\label{table:weights}
\end{table}

%% file: sub_content/table_pyramid.tex
\begin{table}[t]
\centering

\caption{Comparison of different local embedding extractors described in Section~\ref{sec:emd} on 1-shot tasks. $\mathbf{P_{feat}}$ denotes the feature pyramid structure applied on DeepEMD-FCN and $\mathbf{P_{grid}}$ denotes the image pyramid structure applied on DeepEMD-Grid. The parameters in $\mathbf{P_{feat}}$ and $\mathbf{P_{grid}}$ are the RoI pooling size and the grid size, respectively. DeepEMD-Sampling outperforms the plain version of other two methods and both pyramid structures in feature level and image level can effectively boost the performance.}

% \resizebox{0.47\textwidth}{!}
{%
\begin{tabular}{lccc}
\toprule[1pt]
\multicolumn{1}{c}{Method} & Embedding & 5-way& 10-way \\ \hline\hline
\multirow{4}{*}{DeepEMD-FCN}&$5 \times 5$& 65.91  & 49.66         \\ \cline{2-4} 
&$\mathbf{P_{feat}}(5, 3)$  & 66.27  & 49.85         \\ 
&$\mathbf{P_{feat}}(5, 2)$  & 66.42  & 50.02         \\ 
&$\mathbf{P_{feat}}(5, 2, 1)$  & \textbf{66.50}  & \textbf{50.09}         \\ 
\Xhline{0.8pt}
\multirow{9}{*}{DeepEMD-Grid}&$6 \times 6$& 63.61  & 48.72         \\ 
&$5 \times 5$    & 65.38  & 50.08 \\ 
&$4 \times 4$   & 66.57  &  51.42  \\ 
&$3 \times 3$   & 67.31 &	52.26        \\ 
&$2 \times 2$   &66.09 &	50.94 \\ \cline{2-4} 
&$\mathbf{P_{grid}}(3,2 )$    & 67.74 &	52.76 \\ 
&$\mathbf{P_{grid}}(5,3 )$   & 67.01 & 51.89\\ 
&$\mathbf{P_{grid}}(5,2 )$   &  67.39 & 52.20 \\ 
&$\mathbf{P_{grid}}(5,3,2 )$  & \textbf{67.83} & \textbf{52.85} \\ 
\Xhline{0.9pt}
\multirow{3}{*}{DeepEMD-Sampling}&9 patches & 68.09 & 53.04\\ 
& 16 patches & 68.54 & 53.61  
         \\ 
& 25 patches & \textbf{68.77} & \textbf{53.83} 
         \\ 
 \Xhline{1.2pt}
\end{tabular}%
}

\label{table:extractor}
\end{table}

%% file: sub_content/table_ss.tex
\begin{table}[t]
%\small
\centering

\caption{Model pre-training with self-supervised auxiliary tasks. The self-supervised auxiliary task at the pre-training stage can effectively boost the 1-shot performance of our models, while applying the random rotation as data augmentation alone degrades the performance. Please refer to Section~\ref{sec:ablation} for analysis. }

\resizebox{0.47\textwidth}{!}{%
\begin{tabular}{lcccc}
\toprule[1.2pt]
Model & Rotation  &  Self-Supervision & 5-way & 10-way \\ \hline\hline

\multirow{4}{*}{DeepEMD-FCN} & &  & 65.91 & 49.66       \\ 
 &\checkmark & &65.45   & 49.70       \\ 
 & &\checkmark &\textbf{66.64}& \textbf{50.99}  \\\hline
 \multirow{4}{*}{DeepEMD-Grid} & &  & 67.31  & 52.26      \\ 
 &\checkmark & &66.96     & 52.09     \\ 
 & &\checkmark &\textbf{67.87} & \textbf{53.08} \\\hline
 
 \multirow{4}{*}{DeepEMD-Sampling} & &  & 68.77 & 53.83       \\ 
 &\checkmark & &67.98  &    53.42     \\ 
 & &\checkmark &\textbf{69.17} & \textbf{54.52} \\\Xhline{1.2pt}
\end{tabular}%
}

\label{table:ss}
\end{table}

%% file: sub_content/table_cross.tex
\begin{table}[t]

\centering

\caption{Cross-domain experiments (\emph{mini}Imagenet $\rightarrow$ CUB). We report the performance with 95\% confidence intervals on 1-shot 5-way and 5-shot 5-way tasks. We use FCN to extract local features for KNN and our method. Our proposed algorithm outperforms baseline methods with a large margin. }

% \resizebox{0.47\textwidth}{!}
{%
\begin{tabular}{lcc}
\toprule[1.2pt]
\multicolumn{1}{l}{Model}      &  1-shot &5-shot\\ \hline\hline
ProtoNet~\cite{proto}   &50.01 \small{$\pm$ $0.82$}  &72.02 \small{$\pm$ $0.67$}         \\ 
MatchingNet~\cite{matchnet} &51.65 \small{$\pm$ $0.84$}  &69.14 \small{$\pm$ $0.72$}         \\ 
\emph{cosine} classifier~\cite{closer}  &44.17 \small{$\pm$ $0.78$}  &69.01 \small{$\pm$ $0.74$}         \\ 
\emph{linear} classifier~\cite{closer}  &50.37 \small{$\pm$ $0.79$}  &73.30 \small{$\pm$ $0.69$}         \\

KNN~\cite{Revisiting}       &50.84 \small{$\pm$ $0.81$}  &71.25 \small{$\pm$ $0.69$}         \\  \Xhline{1pt}
\textbf{DeepEMD-FCN}  &\textbf{54.24} \small{$\pm$ \textbf{0.86}}  &\textbf{78.86} \small{$\pm$ \textbf{0.65}}         \\
\Xhline{1.2pt}
\end{tabular}%
}

\label{table:cross}
\end{table}

%% file: sub_content/table_time.tex
\begin{table}[t]
%\small
\centering

\caption{Computation time of the EMD layer in DeepEMD-FCN. In a 5-way 1-shot task with 10 query images, we vary the spatial size and the feature dimension of two input feature maps and record their computation time. }

%\resizebox{0.47\textwidth}{!}{%
\begin{tabular}{cccc}
\toprule[1.2pt]
Solver &Spatial Size   &  Dimension & Time\\ \hline\hline

\multirow{12}{*}{QPTH~\cite{amos2017optnet}} & $3 \times 3$& 256 & 0.294 s        \\ 
& $3 \times 3$ &512 &0.294 s        \\ 
 &$3 \times 3$  &1024  &0.295 s         \\ 
  &$3 \times 3$  &2048  &0.295 s         \\ \cline{2-4} 
 &$4 \times 4$& 256 &1.489 s        \\ 
 &$4 \times 4$ &512 &1.493 s         \\ 
&$4 \times 4$  &1024  &1.498 s         \\ 
& $4 \times 4$  &2048  &1.499 s         \\ \cline{2-4} 
& $5 \times 5$& 256 &8.595 s        \\ 
&$5 \times 5$ &512 &8.638 s          \\ 
& $5 \times 5$  &1024  &8.638 s         \\ 
& $5 \times 5$  &2048  &8.650 s  \\\hline
\multirow{4}{*}{OpenCV~\cite{OpenCV}} & $5 \times 5$& 256 &0.018 s        \\ 
 & $5 \times 5$ &512 &0.019 s          \\ 
& $5 \times 5$  &1024  &0.021 s         \\ 
 & $5 \times 5$  &2048  &0.025 s 
      \\\Xhline{1.2pt}
\end{tabular}%
%}

\label{table:time}
\end{table}

%% file: 6_1_soatable.tex
\begin{table*}[t]

	\caption{ Comparison with the state-of-the-art 1-shot 5-way and 5-shot 5-way performance (\textbf{\%}) with 95\% confidence intervals on \textbf{\emph{mini}ImageNet }\textbf{(a)}, \textbf{\emph{tiered}ImageNet} \textbf{(a)}, \textbf{CIFAR-FewShot (a) } \textbf{Fewshot-CIFAR100  }\textbf{(b)}, and \textbf{Caltech-UCSD Birds-200-2011 }\textbf{(c)} datasets. Our model achieves new state-of-the-art performance on all datasets and even outperforms methods with deeper backbones${}^{\dag}$. }

% 	\begin{minipage}{\textwidth}
		\begin{minipage}[t]{1\textwidth}
			%\vspace{20pt}
			\centering
			\begin{subtable}{1\textwidth}
			%\resizebox{\textwidth}{!}{
				\begin{tabular*}{\linewidth}{l @{\extracolsep{\fill}} ccccccc @{}}%{lccccc}
					\toprule[1pt]
					\multicolumn{1}{c}{\multirow{2}{*}{\textbf{Method}}} & \multicolumn{1}{c}{\multirow{2}{*}{\textbf{Backbone}}} & \multicolumn{2}{c}{\textbf{\emph{mini}Imagenet}}                              & \multicolumn{2}{c}{\textbf{\emph{tiered}Imagenet}}             &
					\multicolumn{2}{c}{\textbf{CIFAR-FS}}
					\\ 
					\Xcline{3-8}{1pt}%

					\multicolumn{1}{c}{}                        & \multicolumn{1}{c}{}                          & \multicolumn{1}{c}{\textbf{1-shot}} & \multicolumn{1}{c}{\textbf{5-shot}} &
					\multicolumn{1}{c}{\textbf{1-shot}} & \multicolumn{1}{c}{\textbf{5-shot}} &
					\multicolumn{1}{c}{\textbf{1-shot}} & \multicolumn{1}{c}{\textbf{5-shot}} \\%\midrule[0.5pt]
\Xhline{1pt}

 \textbf{\emph{cosine} classifier~\cite{closer}} & \emph{ResNet12}  & 55.43 \small{$\pm$ $0.81$} & 77.18 \small{$\pm$ $0.61$}  & 61.49 \small{$\pm$ $0.91$}  & 82.37. \small{$\pm$ $0.67$}  & - & -\\

\textbf{TADAM~\cite{TADAM}}  & \emph{ResNet12}  & 58.50 \small{$\pm$ $0.30$} & 76.70 \small{$\pm$ $0.30$}  &  -  &  - & - & - \\

\textbf{ECM~\cite{ECM}}  & \emph{ResNet12}  & 59.00 \small{$\pm$ $-$} & 77.46 \small{$\pm$ $-$}  &  63.99 \small{$\pm$ $-$}  &  81.97 \small{$\pm$ $-$}  & 69.15 \small{$\pm$ $-$} & 84.70 \small{$\pm$ $-$}\\

\textbf{TPN~\cite{tpn}}  & \emph{ResNet12}  & 59.46
 \small{$\pm$ $ -$} & 75.65
 \small{$\pm$ $ -$}  & 59.91 \small{$\pm$ $0.94$}  & 73.30 \small{$\pm$ $0.75$} & - & - \\

\textbf{PPA~\cite{PPA}}  & \emph{WRN\small{-28-10}}${}^{\dag}$  & 59.60 \small{$\pm$ $0.41$} & 73.74  \small{$\pm$ $0.19$}  & 65.65 \small{$\pm$ $0.92$}  & 83.40. \small{$\pm$ $0.65$} & - & - \\

\textbf{Dhillon~\etal~\cite{dhillon2019baseline}}  & \emph{WRN\small{-28-10}}${}^{\dag}$  & 57.73 \small{$\pm$ $0.62$} & 78.17 \small{$\pm$ $0.49$}  & 66.58 \small{$\pm$ $0.70$}  & 85.55 \small{$\pm$ $0.48$} & 68.72 \small{$\pm$ $0.67$} & 86.11 \small{$\pm$ $0.47$} \\

\textbf{ProtoNet~\cite{proto}}  & \emph{ResNet12}  & 60.37 \small{$\pm$ $0.83$} & 78.02 \small{$\pm$ $0.57$}  & 65.65 \small{$\pm$ $0.92$}  & 83.40. \small{$\pm$ $0.65$}  & - & -\\

\textbf{wDAE-GNN~\cite{wDAE}}  & \emph{WRN\small{-28-10}}${}^{\dag}$  & 61.07 \small{$\pm$ $0.15$} & 76.75 \small{$\pm$ $0.11$}  & 68.18 \small{$\pm$ $0.16$}  & 83.09 \small{$\pm$ $0.12$}  & - & -\\

\textbf{MTL~\cite{yaoyao}}  & \emph{ResNet12}  & 61.20 \small{$\pm$ $1.80$} & 75.50 \small{$\pm$ $0.80$}  & -  & -  & - & -\\

\textbf{LEO~\cite{leo}}  & \emph{WRN\small{-28-10}}${}^{\dag}$  & 61.76 \small{$\pm$ $0.08$} & 77.59 \small{$\pm$ $0.12$}  & 66.33 \small{$\pm$ $0.05$}  & 81.44 \small{$\pm$ $0.09$} & - & - \\

\textbf{DC~\cite{dense}}  & \emph{ResNet12}  & 62.53 \small{$\pm$ $0.19$} & 79.77 \small{$\pm$ $0.19$}  & - & - & - & -\\

\textbf{MetaOptNet~\cite{metaoptnet}}  & \emph{ResNet12}  & 62.64 \small{$\pm$ $0.82$} & 78.63 \small{$\pm$ $0.46$}  & 65.99 \small{$\pm$ $0.72$}  & 81.56 \small{$\pm$ $0.53$} & 72.00 \small{$\pm$ $0.70$} & 84.20 \small{$\pm$ $0.50$} \\

\textbf{FEAT~\cite{feat}}  & \emph{ResNet24}${}^{\dag}$  & 62.96 \small{$\pm$ $0.20$} & 78.49 \small{$\pm$ $0.15$}  & - & - & - & - \\

\textbf{MatchNet~\cite{matchnet}}  & \emph{ResNet12}  & 63.08 \small{$\pm$ $0.80$} & 75.99 \small{$\pm$ $0.60$}  & 68.50 \small{$\pm$ $0.92$}  & 80.60 \small{$\pm$ $0.71$} & - & - \\

\textbf{CAN~\cite{CAN}}  & \emph{ResNet12}  & 63.85 \small{$\pm$ $0.48$} & 79.44 \small{$\pm$ $0.34$}  & 69.89 \small{$\pm$ $0.51$}  & 84.23 \small{$\pm$ $0.37$}  & - & -\\

\textbf{DSN~\cite{simon2020adaptive}}  & \emph{ResNet12}  & 62.64 \small{$\pm$ $0.66$} & 78.83 \small{$\pm$ $0.45$}  & 66.22 \small{$\pm$ $0.75$}  & 82.79 \small{$\pm$ $0.48$} & 72.30 \small{$\pm$ $0.80$} & 85.10 \small{$\pm$ $0.60$} \\

\textbf{Tian~\etal~\cite{tian2020rethinking}}  & \emph{ResNet12}  & 64.82 \small{$\pm$ $0.60$} & 82.14 \small{$\pm$ $0.43$}  & 71.52 \small{$\pm$ $0.69$}  & 86.03 \small{$\pm$ $0.49$}  &  73.90 \small{$\pm$ $0.80$} &  86.90 \small{$\pm$ $0.50$}\\

\textbf{CTM~\cite{ctm}}  & \emph{ResNet18}${}^{\dag}$  & 64.12 \small{$\pm$ $0.82$} & 80.51 \small{$\pm$ $0.13$}  & 68.41 \small{$\pm$ $0.39$}  & 84.28 \small{$\pm$ $1.73$} & - & - \\

% \textbf{S2M2-R~\cite{mangla2020charting}}  & \emph{WRN-28-10}${}^{\dag}$  & 64.93 \small{$\pm$ $0.18$} & 83.13 \small{$\pm$ $0.11$}  & 73.71 \small{$\pm$ $0.22$}  & 88.59 \small{$\pm$ $0.14$} & 74.81 \small{$\pm$ $0.19$} & 87.47 \small{$\pm$ $0.13$} \\
\textbf{S2M2-R~\cite{mangla2020charting}}  & \emph{ResNet34}${}^{\dag}$  & 63.74 \small{$\pm$ $0.18$} & 79.45 \small{$\pm$ $0.12$}  & -  & - & 62.77 \small{$\pm$ $0.23$} & 75.75 \small{$\pm$ $0.13$} \\

\textbf{Negative Margin~\cite{liu2020negative}}  & \emph{ResNet12}${}^{\dag}$  & 63.85 \small{$\pm$ $0.81$} & 81.57 \small{$\pm$ $0.56$}  & -  & - &- & - \\

\textbf{Kim~\etal~\cite{kim2020model}}  & \emph{ResNet12}  & 65.08 \small{$\pm$ $0.86$} & 82.70 \small{$\pm$ $0.54$}  & -  & - &  73.51 \small{$\pm$ $0.92$} &  85.49 \small{$\pm$ $0.68$} \\

\textbf{Centroid~\cite{afrasiyabi2020associative}}  & \emph{ResNet18}${}^{\dag}$   & 59.88 \small{$\pm$ $0.67$} & 80.35 \small{$\pm$ $0.73$}  & 69.29 \small{$\pm$ $0.56$}  & 85.97 \small{$\pm$ $0.49$} &  - &  - \\

\textbf{AM3-TADAM~\cite{xing2019adaptive}}  & \emph{ResNet12}  & 65.30 \small{$\pm$ $0.49$} & 78.10 \small{$\pm$ $0.36$}  & 69.08 \small{$\pm$ $0.47$}  & 82.58 \small{$\pm$ $0.31$} &  - &  - \\

\textbf{E$^3$BM~\cite{liu2020ensemble}}  & \emph{ResNet25}${}^{\dag}$   & 64.30 \small{$\pm$ -} & 81.00 \small{$\pm$ -}  & 70.00 \small{$\pm$ -}  & 85.00 \small{$\pm$ -} &  - &  - \\

\Xhline{1pt}

\textbf{DeepEMD-FCN}  & \emph{ResNet12}  & \textbf{66.50} \small{$\pm$ \textbf{0.80}} & \textbf{82.41} \small{$\pm$ \textbf{0.56}}  & \textbf{72.65} \small{$\pm$ \textbf{0.31}}  & \textbf{86.03} \small{$\pm$ \textbf{0.58}}  & \textbf{74.58} \small{$\pm$ \textbf{0.29}}& \textbf{86.92} \small{$\pm$ \textbf{0.41}}\\ 

\textbf{DeepEMD-Grid}  & \emph{ResNet12}  & \textbf{67.83} \small{$\pm$ \textbf{0.29}} & \textbf{83.14} \small{$\pm$ \textbf{0.57}}  & \textbf{73.13} \small{$\pm$ \textbf{0.32}}  & \textbf{87.08} \small{$\pm$ \textbf{0.60}} & \textbf{73.31} \small{$\pm$ \textbf{0.29}}& \textbf{85.43} \small{$\pm$ \textbf{0.37}}\\ 

\textbf{DeepEMD-Sampling}  & \emph{ResNet12}  & \textbf{68.77} \small{$\pm$ \textbf{0.29}} & \textbf{84.13} \small{$\pm$ \textbf{0.53}}  & \textbf{74.29} \small{$\pm$ \textbf{0.32}}  & \textbf{86.98} \small{$\pm$ \textbf{0.60}} &\textbf{74.48} \small{$\pm$ \textbf{0.29}}&\textbf{86.37} \small{$\pm$ \textbf{0.36}}\\ 
\Xhline{1pt}

				\end{tabular*}
				
	%	}
		\caption{ Results on \textbf{\emph{mini}ImageNet}, \textbf{\emph{tiered}ImageNet}, and \textbf{CIFAR-FewShot datasets}. }
		\end{subtable}
		
		\end{minipage}

		\begin{minipage}{1\textwidth}
		\begin{minipage}[t]{0.5\textwidth}
		%	\vspace{0pt}
\centering
			\begin{subtable}{1\textwidth}
				\resizebox{\textwidth}{!}{
					
					\begin{tabular}{lccc}
						\toprule[1pt]
						\multicolumn{1}{c}{\textbf{Method}} &\textbf{ Backbone} & \textbf{1-shot}        & \textbf{5-shot }       \\ \Xhline{1pt}
\textbf{\emph{cosine} classifier~\cite{closer}} & \emph{ResNet12} & 38.47 \small{$\pm$ $0.70$} & 57.67 \small{$\pm$ $0.77$} \\

\textbf{TADAM~\cite{TADAM}} & \emph{ResNet12} & 40.10 \small{$\pm$ $0.40$} & 56.10 \small{$\pm$ $0.40$} \\

\textbf{MetaOptNet~\cite{metaoptnet}} & \emph{ResNet12} &  41.10\small{$\pm$ $0.60$} & 55.5 \small{$\pm$ $0.60$} \\

\textbf{ProtoNet~\cite{proto}} & \emph{ResNet12} & 41.54 \small{$\pm$ $0.76$} & 57.08 \small{$\pm$ $0.76$} \\

\textbf{DC~\cite{dense}} & \emph{ResNet12} & 42.04 \small{$\pm$ $0.17$} & 57.05 \small{$\pm$ $0.16$} \\

\textbf{MatchNet~\cite{matchnet}} & \emph{ResNet12} & 43.88 \small{$\pm$ $0.75$} & 57.05 \small{$\pm$ $0.71$} \\

\textbf{MTL~\cite{yaoyao}} & \emph{ResNet12} &  45.10\small{$\pm$ $1.8$} & 57.6 \small{$\pm$ $0.9$} \\

\textbf{Centroid~\cite{afrasiyabi2020associative}}  & \emph{ResNet18}${}^{\dag}$  &  45.83\small{$\pm$ $0.48$} & 59.74 \small{$\pm$ $0.56$} \\

\textbf{Tian~\etal~\cite{tian2020rethinking}}  & \emph{ResNet12}   &  44.60\small{$\pm$ $0.70$} & 60.90 \small{$\pm$ $0.6$} \\

\textbf{E$^3$BM~\cite{liu2020ensemble}}  & \emph{ResNet25}${}^{\dag}$   &  45.00\small{$\pm$ -} & 60.50 \small{$\pm$ -} \\

\Xhline{1pt}

\textbf{DeepEMD-FCN} & \emph{ResNet12} &  \textbf{46.60} \small{$\pm$ \textbf{0.26}} &  \textbf{63.22} \small{$\pm$ \textbf{0.71}} \\

\textbf{DeepEMD-Grid} & \emph{ResNet12} &  \textbf{45.23} \small{$\pm$ \textbf{0.26}} &  \textbf{61.39} \small{$\pm$ \textbf{0.76}} \\

\textbf{DeepEMD-Sampling} & \emph{ResNet12} &  \textbf{45.37} \small{$\pm$ \textbf{0.25}} &  \textbf{61.51} \small{$\pm$ \textbf{0.70}} \\\bottomrule[1pt]

					\end{tabular}
				}
				\caption{Results on \textbf{Fewshot-CIFAR100} dataset.}
			\end{subtable}
			\end{minipage}
			\begin{minipage}[t]{0.5\textwidth}
			\begin{subtable}{1\textwidth}
				\resizebox{\textwidth}{!}{
					\begin{tabular}{lccc}
						\toprule[1pt]
						
				\multicolumn{1}{c}{\textbf{Method}} &\textbf{ Backbone} & \textbf{1-shot}        & \textbf{5-shot }       \\ \Xhline{1pt}		
\textbf{ProtoNet~\cite{proto}} & \emph{ResNet12} & 66.09 \small{$\pm$ $0.92$} & 82.50 \small{$\pm$ $0.58$} \\	

\textbf{RelationNet~\cite{relation,closer}} & \emph{ResNet34}${}^{\dag}$ & 66.20 \small{$\pm$ $0.99$} & 82.30 \small{$\pm$ $0.58$} \\						

\textbf{DEML~\cite{Zhou2018deep}} & \emph{ResNet50}${}^{\dag}$ & 66.95 \small{$\pm$ $1.06$} & 77.11 \small{$\pm$ $0.78$} \\

\textbf{MAML~\cite{maml,closer}} & \emph{ResNet34}${}^{\dag}$ & 67.28 \small{$\pm$ $1.08$} & 83.47 \small{$\pm$ $0.59$} \\

\textbf{\emph{cosine} classifier~\cite{closer}} & \emph{ResNet12} &  67.30 \small{$\pm$ $0.86$} & 84.75 \small{$\pm$ $0.60$} \\

\textbf{MatchNet~\cite{matchnet}} & \emph{ResNet12} & 71.87 \small{$\pm$ $0.85$} & 85.08 \small{$\pm$ $0.57$} \\

% \textbf{Negative Margin~\cite{liu2020negative}} & \emph{ResNet12}${}^{\dag}$ & 72.66 \small{$\pm$ $0.85$} & 89.40 \small{$\pm$ $0.43$} \\

\textbf{S2M2-R~\cite{mangla2020charting}} & \emph{ResNet34}${}^{\dag}$ & 72.92 \small{$\pm$ $0.83$} & 86.55 \small{$\pm$ $0.51$} \\

\textbf{Centroid Align~\cite{afrasiyabi2020associative}}  & \emph{ResNet18}${}^{\dag}$  &  74.22\small{$\pm$ $1.09$} & 88.65 \small{$\pm$ $0.55$} \\

\textbf{Adversarial Align~\cite{afrasiyabi2020associative}}  & \emph{ResNet18}${}^{\dag}$  &  73.87\small{$\pm$ $0.76$} & 84.95 \small{$\pm$ $0.59$} \\

% \textbf{In. SVM~\cite{wang2020instance}} & \emph{ResNet12} & 75.84 \small{$\pm$ $-$}~~~ & 89.26 \small{$\pm$ $-$}~~~  \\

\Xhline{1pt}

\textbf{DeepEMD-FCN} & \emph{ResNet12} &  \textbf{77.14} \small{$\pm$ \textbf{0.29}} &  \textbf{88.98} \small{$\pm$ \textbf{0.49}} \\

\textbf{DeepEMD-Grid} & \emph{ResNet12} &  \textbf{77.64} \small{$\pm$ \textbf{0.29}} &  \textbf{89.25} \small{$\pm$ \textbf{0.53}} \\

\textbf{DeepEMD-Sampling} & \emph{ResNet12} &  \textbf{79.27} \small{$\pm$ \textbf{0.29}} &  \textbf{89.80} \small{$\pm$ \textbf{0.51}} \\
					\Xhline{1pt}	
					\end{tabular}
				}
			\caption{Results on the 
			\textbf{Caltech-UCSD Birds-200-2011} dataset.}
			\end{subtable}
		\end{minipage}

	\label{table:soa}	
 	\end{minipage}
\end{table*}

%% file: sub_content/table_metric.tex
\begin{table}[t]
\centering

\caption{Image retrieval experiment on Caltech-UCSD  Birds-200-2011 dataset. Our method can effectively improve the performance of deep metric learning methods. The relative performance improvements over the baselines are indicated ($\uparrow$).}

 \resizebox{0.47\textwidth}{!}
{%
\begin{tabular}{llccc}
\toprule[1pt]
Method & \multicolumn{1}{c}{Method} & \textbf{P@1}~$\uparrow$ & \textbf{RP}~$\uparrow$ & \textbf{MAP@R}~$\uparrow$\\ \hline\hline

% \multirow{4}{*}{Pretrain Only}& Raw & 50.30 ~~~~~~~~~~ & 22.78 ~~~~~~~~~ & 12.60   ~~~~~~~~~   \\ 
% & ~+DeepEMD-F  &  55.76 \tiny{$\uparrow 5.46$} & 26.18 \tiny{$\uparrow 3.40$} & 15.46 \tiny{$\uparrow 2.86$}  \\ 
% & ~+DeepEMD-G  &  52.40 \tiny{$\uparrow 2.10$}~ & 23.40 \tiny{$\uparrow 0.62$} & 13.09 \tiny{$\uparrow 0.49$}  \\ 
% & ~+DeepEMD-S  &  54.24 \tiny{$\uparrow 3.94 $} & 23.57 \tiny{$\uparrow 0.79$} & 13.49 \tiny{$\uparrow 0.89$}    \\ \hline

\multirow{4}{*}{N. Softmax~\cite{zhai2018classification}}& Raw & 61.83 ~~~~~~~~ & 33.18 ~~~~~~~~ & 22.18   ~~~~~~~~   \\ 
& ~+DeepEMD-F  & 63.40 \tiny{$\uparrow 1.57$} & 34.41 \tiny{$\uparrow 1.23$}  &    23.42 \tiny{$\uparrow 1.24$}  \\ 
& ~+DeepEMD-G  & 63.93 \tiny{$\uparrow 2.10$}  & 34.27 \tiny{$\uparrow 1.09$} &    23.38 \tiny{$\uparrow 1.20$}  \\ 
& ~+DeepEMD-S  &  64.12 \tiny{$\uparrow 2.29$}&35.40 \tiny{$\uparrow 2.22$}    & 24.34 \tiny{$\uparrow 2.16$}    \\ \hline

\multirow{4}{*}{Contrastive~\cite{hadsell2006dimensionality}}& Raw & 62.51 ~~~~~~~~ & 34.68 ~~~~~~~~  & 23.58  ~~~~~~~~    \\ 
& ~+DeepEMD-F  & 64.74 \tiny{$\uparrow 2.23$} & 35.77 \tiny{$\uparrow 1.09$}  &    24.78 \tiny{$\uparrow 1.20$}  \\ 
& ~+DeepEMD-G  & 66.58 \tiny{$\uparrow 4.07$} & 36.85 \tiny{$\uparrow 2.17$} & 25.79    \tiny{$\uparrow 2.21$}  \\ 
& ~+DeepEMD-S  & 67.74 \tiny{$\uparrow 5.23$} & 37.91 \tiny{$\uparrow 3.23$}   &     26.78 \tiny{$\uparrow 3.20$}\\ \hline

\multirow{4}{*}{SoftTriple~\cite{qian2019softtriple}}& Raw & 62.05 ~~~~~~~~ & 33.98 ~~~~~~~~ &   23.34 ~~~~~~~~   \\  
& ~+DeepEMD-F  & 63.15 \tiny{$\uparrow 1.10$} & 34.82 \tiny{$\uparrow 0.84$} &   24.19    \tiny{$\uparrow 0.85$}\\ 
& ~+DeepEMD-G  & 63.57 \tiny{$\uparrow 1.52$} & 35.08 \tiny{$\uparrow 1.10$}&    24.56 \tiny{$\uparrow 1.22$}  \\ 
& ~+DeepEMD-S  &  65.19 \tiny{$\uparrow 3.14$}& 35.90  \tiny{$\uparrow 1.92$ }&    25.33 \tiny{$\uparrow 1.99$ } \\ \hline

\multirow{4}{*}{SNR~\cite{yuan2019signal}}& Raw & 63.63 ~~~~~~~~ & 34.48 ~~~~~~~~ & 23.45   ~~~~~~~~   \\ 
& ~+DeepEMD-F  & 65.21 \tiny{$\uparrow 1.58$} & 35.71 \tiny{$\uparrow 1.23$}  &    24.80 \tiny{$\uparrow 1.35$}  \\ 
& ~+DeepEMD-G  & 67.74 \tiny{$\uparrow 4.11$}  & 36.74 \tiny{$\uparrow 2.26$} &    25.78 \tiny{$\uparrow 2.33$}  \\ 
& ~+DeepEMD-S  &  68.52 \tiny{$\uparrow 4.89$}& 37.90 \tiny{$\uparrow 3.42$}    & 26.91 \tiny{$\uparrow 3.46$}    \\ \hline

\multirow{4}{*}{ArcFace~\cite{deng2019arcface}}& Raw & 62.74 ~~~~~~~~ & 34.01 ~~~~~~~~ & 22.85  ~~~~~~~~    \\ 
& ~+DeepEMD-F  & 63.92 \tiny{$\uparrow 1.45$} & 35.21 \tiny{$\uparrow 1.20$} &    24.10 \tiny{$\uparrow 1.25$}  \\ 
& ~+DeepEMD-G  & 65.29 \tiny{$\uparrow 2.55$}  & 35.56 \tiny{$\uparrow 1.55$}&   24.58  \tiny{$\uparrow 1.73$}  \\ 
& ~+DeepEMD-S  & 66.49 \tiny{$\uparrow 3.75$}  & 36.82 \tiny{$\uparrow 2.81$}  &  25.67 \tiny{$\uparrow 2.82$}   \\ \hline

\multirow{4}{*}{MS~\cite{wang2019multi}}& Raw & 63.99 ~~~~~~~~ &  33.71 ~~~~~~~~ &   22.73 ~~~~~~~~   \\  
& ~+DeepEMD-F  & 66.39 \tiny{$\uparrow 2.40$} & 35.82 \tiny{$\uparrow 2.11$} &  24.89 \tiny{$\uparrow 2.16$}    \\ 
& ~+DeepEMD-G  &  67.94 \tiny{$\uparrow 3.95$}& 36.60 \tiny{$\uparrow 2.89$}&  25.65 \tiny{$\uparrow 2.92$}     \\  
& ~+DeepEMD-S  &  69.51 \tiny{$\uparrow 5.52$} &  38.29 \tiny{$\uparrow 4.58$} &    27.38 \tiny{$\uparrow 4.65$} \\  

 \Xhline{1.2pt}
\end{tabular}%
}

\label{table:metric}
\end{table}

%% file: 6_conclusion.tex
\section{Conclusion}

We have proposed a few-shot classification framework that employs the 
Earth Mover's Distance as the distance metric.  The implicit function theorem allows our network to be end-to-end trainable. Our proposed cross-reference mechanism for setting the weights of nodes turns out crucial in the EMD formulation and can effectively minimize the negative impact caused by irrelevant regions. The learnable structured fully connected layer can directly classify dense representations of images in the $k$-shot settings. Our algorithm achieves new state-of-the-art performance on multiple datasets.

% \section*{Acknowledgements} 

% \appendices

% % you can choose not to have a title for an appendix
% % if you want by leaving the argument blank
% \section{}

\ifCLASSOPTIONcompsoc
  \section*{Acknowledgments}
\else
  \section*{Acknowledgment}
\fi

This work was supported by National Key R\&D Program of China (No. 2022ZD0118700).

This research is supported by the National Research Foundation, Singapore under its AI Singapore Programme (AISG Award No: AISG-RP-2018-003), and the MOE AcRF Tier-1 research grants: RG95/20.

%% file: bare_jrnl_compsoc.bbl
\begin{thebibliography}{100}
\providecommand{\url}[1]{#1}
\csname url@samestyle\endcsname
\providecommand{\newblock}{\relax}
\providecommand{\bibinfo}[2]{#2}
\providecommand{\BIBentrySTDinterwordspacing}{\spaceskip=0pt\relax}
\providecommand{\BIBentryALTinterwordstretchfactor}{4}
\providecommand{\BIBentryALTinterwordspacing}{\spaceskip=\fontdimen2\font plus
\BIBentryALTinterwordstretchfactor\fontdimen3\font minus
  \fontdimen4\font\relax}
\providecommand{\BIBforeignlanguage}[2]{{%
\expandafter\ifx\csname l@#1\endcsname\relax
\typeout{** WARNING: IEEEtran.bst: No hyphenation pattern has been}%
\typeout{** loaded for the language `#1'. Using the pattern for}%
\typeout{** the default language instead.}%
\else
\language=\csname l@#1\endcsname
\fi
#2}}
\providecommand{\BIBdecl}{\relax}
\BIBdecl

\bibitem{matchnet}
O.~Vinyals, C.~Blundell, T.~Lillicrap, K.~Kavukcuoglu, and D.~Wierstra,
  ``Matching networks for one shot learning,'' in \emph{Proc. Advances in
  Neural Inf. Process. Syst.}, 2016.

\bibitem{proto}
J.~Snell, K.~Swersky, and R.~S. Zemel, ``Prototypical networks for few-shot
  learning,'' in \emph{Proc. Advances in Neural Inf. Process. Syst.}, 2017.

\bibitem{feat}
H.~Ye, H.~Hu, D.~Zhan, and F.~Sha, ``Learning embedding adaptation for few-shot
  learning,'' \emph{arXiv}, vol. 1812.03664, 2018.

\bibitem{TADAM}
B.~N. Oreshkin, P.~Rodr{\'{\i}}guez, and A.~Lacoste, ``{TADAM:} task dependent
  adaptive metric for improved few-shot learning,'' in \emph{Proc. Advances in
  Neural Inf. Process. Syst.}, 2018.

\bibitem{relation}
F.~Sung, Y.~Yang, L.~Zhang, T.~Xiang, P.~H.~S. Torr, and T.~M. Hospedales,
  ``Learning to compare: Relation network for few-shot learning,'' in
  \emph{Proc. IEEE Conf. Comp. Vis. Patt. Recogn.}, 2018.

\bibitem{Revisiting}
W.~Li, L.~Wang, J.~Xu, J.~Huo, Y.~Gao, and J.~Luo, ``Revisiting local
  descriptor based image-to-class measure for few-shot learning,'' in
  \emph{Proc. IEEE Conf. Comp. Vis. Patt. Recogn.}, June 2019.

\bibitem{cnn}
A.~Krizhevsky, I.~Sutskever, and G.~E. Hinton, ``Imagenet classification with
  deep convolutional neural networks,'' in \emph{Proc. Advances in Neural Inf.
  Process. Syst.}, 2012, pp. 1097--1105.

\bibitem{resnet}
K.~He, X.~Zhang, S.~Ren, and J.~Sun, ``Deep residual learning for image
  recognition,'' in \emph{Proc. IEEE Conf. Comp. Vis. Patt. Recogn.}, 2016, pp.
  770--778.

\bibitem{emd}
Y.~Rubner, C.~Tomasi, and L.~J. Guibas, ``The earth mover's distance as a
  metric for image retrieval,'' \emph{Int. J. Comput. Vision}, vol.~40, no.~2,
  pp. 99--121, 2000.

\bibitem{transportation}
F.~L. Hitchcock, ``The distribution of a product from several sources to
  numerous localities,'' \emph{J. Mathematics and Physics}, vol.~20, no. 1-4,
  pp. 224--230, 1941.

\bibitem{krantz2012implicit}
S.~G. Krantz and H.~R. Parks, \emph{The implicit function theorem: history,
  theory, and applications}.\hskip 1em plus 0.5em minus 0.4em\relax Springer
  Science \& Business Media, 2012.

\bibitem{dontchev2009implicit}
A.~L. Dontchev and R.~T. Rockafellar, ``Implicit functions and solution
  mappings,'' \emph{Springer Monographs in Mathematics. Springer}, vol. 208,
  2009.

\bibitem{barratt2018differentiability}
S.~Barratt, ``On the differentiability of the solution to convex optimization
  problems,'' \emph{arXiv preprint arXiv:1804.05098}, 2018.

\bibitem{zhang2020deepemd}
C.~Zhang, Y.~Cai, G.~Lin, and C.~Shen, ``{DeepEMD}: Few-shot image
  classification with differentiable earth mover's distance and structured
  classifiers,'' in \emph{Proc. IEEE Conf. Comp. Vis. Patt. Recogn.}, 2020.

\bibitem{Keshari18}
R.~Keshari, M.~Vatsa, R.~Singh, and A.~Noore, ``Learning structure and strength
  of {CNN} filters for small sample size training,'' in \emph{Proc. IEEE Conf.
  Comp. Vis. Patt. Recogn.}, 2018.

\bibitem{FeiFeiFP06}
F.~Li, R.~Fergus, and P.~Perona, ``One-shot learning of object categories,''
  \emph{{IEEE} Trans. Pattern Anal. Mach. Intell.}, vol.~28, no.~4, pp.
  594--611, 2006.

\bibitem{GidarisCVPR2018}
S.~Gidaris and N.~Komodakis, ``Dynamic few-shot visual learning without
  forgetting,'' in \emph{Proc. IEEE Conf. Comp. Vis. Patt. Recogn.}, 2018.

\bibitem{YanZH19}
S.~Yan, S.~Zhang, and X.~He, ``A dual attention network with semantic embedding
  for few-shot learning,'' in \emph{Proc. {AAAI} Conf. Artificial Intell.},
  2019.

\bibitem{Wertheimer_2019_CVPR}
D.~Wertheimer and B.~Hariharan, ``Few-shot learning with localization in
  realistic settings,'' in \emph{Proc. IEEE Conf. Comp. Vis. Patt. Recogn.},
  June 2019.

\bibitem{wang2020instance}
Y.~Wang, C.~Xu, C.~Liu, L.~Zhang, and Y.~Fu, ``Instance credibility inference
  for few-shot learning,'' in \emph{Proc. IEEE Conf. Comp. Vis. Patt. Recogn.},
  2020, pp. 12\,836--12\,845.

\bibitem{afrasiyabi2020associative}
A.~Afrasiyabi, J.-F. Lalonde, and C.~Gagn{\'e}, ``Associative alignment for
  few-shot image classification,'' in \emph{Proc. Eur. Conf. Comp. Vis.}\hskip
  1em plus 0.5em minus 0.4em\relax Springer, 2020, pp. 18--35.

\bibitem{liu2020negative}
B.~Liu, Y.~Cao, Y.~Lin, Q.~Li, Z.~Zhang, M.~Long, and H.~Hu, ``Negative margin
  matters: Understanding margin in few-shot classification,'' in \emph{Proc.
  Eur. Conf. Comp. Vis.}\hskip 1em plus 0.5em minus 0.4em\relax Springer, 2020,
  pp. 438--455.

\bibitem{lichtenstein2020tafssl}
M.~Lichtenstein, P.~Sattigeri, R.~Feris, R.~Giryes, and L.~Karlinsky, ``Tafssl:
  Task-adaptive feature sub-space learning for few-shot classification,'' in
  \emph{Proc. Eur. Conf. Comp. Vis.}\hskip 1em plus 0.5em minus 0.4em\relax
  Springer, 2020, pp. 522--539.

\bibitem{yu2020transmatch}
Z.~Yu, L.~Chen, Z.~Cheng, and J.~Luo, ``Transmatch: A transfer-learning scheme
  for semi-supervised few-shot learning,'' in \emph{Proc. IEEE Conf. Comp. Vis.
  Patt. Recogn.}, 2020, pp. 12\,856--12\,864.

\bibitem{xuattentional}
W.~Xu, Y.~Xu, H.~Wang, and Z.~Tu, ``Attentional constellation nets for few-shot
  learning,'' 2021.

\bibitem{MunkhdalaiICML18}
T.~Munkhdalai, X.~Yuan, S.~Mehri, and A.~Trischler, ``Rapid adaptation with
  conditionally shifted neurons,'' in \emph{Proc. Int. Conf. Mach. Learn.},
  2018.

\bibitem{SantoroBBWL16}
A.~Santoro, S.~Bartunov, M.~Botvinick, D.~Wierstra, and T.~P. Lillicrap,
  ``Meta-learning with memory-augmented neural networks,'' in \emph{Proc. Int.
  Conf. Mach. Learn.}, 2016.

\bibitem{closer}
W.-Y. Chen, Y.-C. Liu, Z.~Kira, Y.-C. Wang, and J.-B. Huang, ``A closer look at
  few-shot classification,'' in \emph{Proc. Int. Conf. Learn. Representations},
  2019.

\bibitem{MunkhdalaiICML2017}
T.~Munkhdalai and H.~Yu, ``Meta networks,'' in \emph{Proc. Int. Conf. Mach.
  Learn.}, 2017.

\bibitem{ScottNIPS2018}
T.~R. Scott, K.~Ridgeway, and M.~C. Mozer, ``Adapted deep embeddings: {A}
  synthesis of methods for k-shot inductive transfer learning,'' in \emph{Proc.
  Advances in Neural Inf. Process. Syst.}, 2018.

\bibitem{PPA}
S.~Qiao, C.~Liu, W.~Shen, and A.~L. Yuille, ``Few-shot image recognition by
  predicting parameters from activations,'' in \emph{Proc. IEEE Conf. Comp.
  Vis. Patt. Recogn.}, 2018.

\bibitem{Zhou2018deep}
F.~Zhou, B.~Wu, and Z.~Li, ``Deep meta-learning: Learning to learn in the
  concept space,'' \emph{arXiv}, vol. 1802.03596, 2018.

\bibitem{MetzICLR19}
L.~Metz, N.~Maheswaranathan, B.~Cheung, and J.~Sohl-Dickstein, ``Meta-learning
  update rules for unsupervised representation learning,'' in \emph{Proc. Int.
  Conf. Learn. Representations}, 2019.

\bibitem{LeeICML18}
Y.~Lee and S.~Choi, ``Gradient-based meta-learning with learned layerwise
  metric and subspace,'' in \emph{Proc. Int. Conf. Mach. Learn.}, 2018.

\bibitem{LiICML2018}
Z.~Li, F.~Zhou, F.~Chen, and H.~Li, ``Meta-sgd: Learning to learn quickly for
  few shot learning,'' in \emph{Proc. Int. Conf. Mach. Learn.}, 2018.

\bibitem{Luketina2016}
J.~Luketina, T.~Raiko, M.~Berglund, and K.~Greff, ``Scalable gradient-based
  tuning of continuous regularization hyperparameters,'' in \emph{Proc. Int.
  Conf. Mach. Learn.}, 2016.

\bibitem{Naik92}
D.~K. Naik and R.~Mammone, ``Meta-neural networks that learn by learning,'' in
  \emph{Proc. Int. Joint Conf. Neural Networks}, 1992.

\bibitem{dense}
Y.~Lifchitz, Y.~Avrithis, S.~Picard, and A.~Bursuc, ``Dense classification and
  implanting for few-shot learning,'' in \emph{Proc. IEEE Conf. Comp. Vis.
  Patt. Recogn.}, June 2019.

\bibitem{Schonfeld_2019_CVPR}
E.~Schonfeld, S.~Ebrahimi, S.~Sinha, T.~Darrell, and Z.~Akata, ``Generalized
  zero- and few-shot learning via aligned variational autoencoders,'' in
  \emph{Proc. IEEE Conf. Comp. Vis. Patt. Recogn.}, June 2019.

\bibitem{flennerhag2019meta}
S.~Flennerhag, A.~A. Rusu, R.~Pascanu, H.~Yin, and R.~Hadsell, ``Meta-learning
  with warped gradient descent,'' \emph{arXiv preprint arXiv:1909.00025}, 2019.

\bibitem{li2019learning}
X.~Li, Q.~Sun, Y.~Liu, Q.~Zhou, S.~Zheng, T.-S. Chua, and B.~Schiele,
  ``Learning to self-train for semi-supervised few-shot classification,'' in
  \emph{Proc. Advances in Neural Inf. Process. Syst.}, 2019, pp.
  10\,276--10\,286.

\bibitem{park2019meta}
E.~Park and J.~B. Oliva, ``Meta-curvature,'' in \emph{Proc. Advances in Neural
  Inf. Process. Syst.}, 2019, pp. 3309--3319.

\bibitem{franceschi2018bilevel}
L.~Franceschi, P.~Frasconi, S.~Salzo, R.~Grazzi, and M.~Pontil, ``Bilevel
  programming for hyperparameter optimization and meta-learning,'' \emph{arXiv
  preprint arXiv:1806.04910}, 2018.

\bibitem{rajeswaran2019meta}
A.~Rajeswaran, C.~Finn, S.~M. Kakade, and S.~Levine, ``Meta-learning with
  implicit gradients,'' in \emph{Proc. Advances in Neural Inf. Process. Syst.},
  2019, pp. 113--124.

\bibitem{FranceschiICML18}
L.~Franceschi, P.~Frasconi, S.~Salzo, R.~Grazzi, and M.~Pontil, ``Bilevel
  programming for hyperparameter optimization and meta-learning,'' in
  \emph{Proc. Int. Conf. Mach. Learn.}, 2018.

\bibitem{qi2018low}
H.~Qi, M.~Brown, and D.~G. Lowe, ``Low-shot learning with imprinted weights,''
  in \emph{Proc. IEEE Conf. Comp. Vis. Patt. Recogn.}, 2018, pp. 5822--5830.

\bibitem{simon2020adaptive}
C.~Simon, P.~Koniusz, R.~Nock, and M.~Harandi, ``Adaptive subspaces for
  few-shot learning,'' in \emph{Proc. IEEE Conf. Comp. Vis. Patt. Recogn.},
  2020, pp. 4136--4145.

\bibitem{kim2020model}
J.~Kim, H.~Kim, and G.~Kim, ``Model-agnostic boundary-adversarial sampling for
  test-time generalization in few-shot learning,'' \emph{Proc. Eur. Conf. Comp.
  Vis.}, pp. 599--617, 2020.

\bibitem{zhangiept}
M.~Zhang, J.~Zhang, Z.~Lu, T.~Xiang, M.~Ding, and S.~Huang, ``{IEPT}:
  Instance-level and episode-level pre-text tasks for few-shot learning,''
  2021.

\bibitem{fei2021melr}
N.~Fei, Z.~Lu, T.~Xiang, and S.~Huang, ``Melr: Meta-learning via modeling
  episode-level relationships for few-shot learning,'' in \emph{Proc. Int.
  Conf. Learn. Representations}, 2021.

\bibitem{oh2021boil}
J.~Oh, H.~Yoo, C.~Kim, and S.-Y. Yun, ``{BOIL}: Towards representation change
  for few-shot learning,'' in \emph{Proc. Int. Conf. Learn. Representations},
  2021.

\bibitem{snell2020bayesian}
J.~Snell and R.~Zemel, ``{B}ayesian few-shot classification with one-vs-each
  polya-gamma augmented {G}aussian processes,'' in \emph{Proc. Int. Conf.
  Learn. Representations}, 2021.

\bibitem{patacchiola2020bayesian}
M.~Patacchiola, J.~Turner, E.~J. Crowley, M.~O'Boyle, and A.~Storkey,
  ``Bayesian meta-learning for the few-shot setting via deep kernels,'' 2020.

\bibitem{maml}
C.~Finn, P.~Abbeel, and S.~Levine, ``Model-agnostic meta-learning for fast
  adaptation of deep networks,'' in \emph{Proc. Int. Conf. Mach. Learn.}, 2017.

\bibitem{trainmaml}
A.~Antoniou, H.~Edwards, and A.~Storkey, ``How to train your maml,'' in
  \emph{Proc. Int. Conf. Learn. Representations}, 2019.

\bibitem{leo}
A.~A. Rusu, D.~Rao, J.~Sygnowski, O.~Vinyals, R.~Pascanu, S.~Osindero, and
  R.~Hadsell, ``Meta-learning with latent embedding optimization,'' in
  \emph{Proc. Int. Conf. Learn. Representations}, 2019.

\bibitem{yaoyao}
Q.~Sun, Y.~Liu, T.-S. Chua, and B.~Schiele, ``Meta-transfer learning for
  few-shot learning,'' in \emph{Proc. IEEE Conf. Comp. Vis. Patt. Recogn.},
  2019.

\bibitem{Jamal_2019_CVPR}
M.~A. Jamal and G.-J. Qi, ``Task agnostic meta-learning for few-shot
  learning,'' in \emph{Proc. IEEE Conf. Comp. Vis. Patt. Recogn.}, June 2019.

\bibitem{CAN}
R.~Hou, H.~Chang, B.~Ma, S.~Shan, and X.~Chen, ``Cross attention network for
  few-shot classification,'' in \emph{Proc. Advances in Neural Inf. Process.
  Syst.}, 2019.

\bibitem{ctm}
H.~Li, D.~Eigen, S.~Dodge, M.~Zeiler, and X.~Wang, ``Finding task-relevant
  features for few-shot learning by category traversal,'' in \emph{Proc. IEEE
  Conf. Comp. Vis. Patt. Recogn.}, 2019.

\bibitem{tieredimagenet}
M.~Ren, E.~Triantafillou, S.~Ravi, J.~Snell, K.~Swersky, J.~B. Tenenbaum,
  H.~Larochelle, and R.~S. Zemel, ``Meta-learning for semi-supervised few-shot
  classification,'' in \emph{Proc. Int. Conf. Learn. Representations}, 2018.

\bibitem{liu2019prototype}
J.~Liu, L.~Song, and Y.~Qin, ``Prototype rectification for few-shot learning,''
  \emph{arXiv preprint arXiv:1911.10713}, 2019.

\bibitem{xing2019adaptive}
C.~Xing, N.~Rostamzadeh, B.~N. Oreshkin, and P.~O. Pinheiro, ``Adaptive
  cross-modal few-shot learning,'' \emph{arXiv preprint arXiv:1902.07104},
  2019.

\bibitem{Zhang_2021_CVPR}
C.~Zhang, N.~Song, G.~Lin, Y.~Zheng, P.~Pan, and Y.~Xu, ``Few-shot incremental
  learning with continually evolved classifiers,'' in \emph{Proc. IEEE Conf.
  Comp. Vis. Patt. Recogn.}, June 2021.

\bibitem{crosstrans}
C.~Doersch, A.~Gupta, and A.~Zisserman, ``{CrossTransformers}: spatially-aware
  few-shot transfer,'' in \emph{Proc. Advances in Neural Inf. Process. Syst.},
  2020.

\bibitem{Kim_2019_CVPR}
J.~Kim, T.~Kim, S.~Kim, and C.~D. Yoo, ``Edge-labeling graph neural network for
  few-shot learning,'' in \emph{Proc. IEEE Conf. Comp. Vis. Patt. Recogn.},
  June 2019.

\bibitem{wDAE}
S.~Gidaris and N.~Komodakis, ``Generating classification weights with gnn
  denoising autoencoders for few-shot learning,'' in \emph{Proc. IEEE Conf.
  Comp. Vis. Patt. Recogn.}, June 2019.

\bibitem{graph}
V.~Garcia and J.~Bruna, ``Few-shot learning with graph neural networks,''
  \emph{arXiv preprint arXiv:1711.04043}, 2017.

\bibitem{yang2020dpgn}
L.~Yang, L.~Li, Z.~Zhang, X.~Zhou, E.~Zhou, and Y.~Liu, ``Dpgn: Distribution
  propagation graph network for few-shot learning,'' in \emph{Proc. IEEE Conf.
  Comp. Vis. Patt. Recogn.}, 2020, pp. 13\,390--13\,399.

\bibitem{rl}
W.-H. Chu, Y.-J. Li, J.-C. Chang, and Y.-C.~F. Wang, ``Spot and learn: A
  maximum-entropy patch sampler for few-shot image classification,'' in
  \emph{Proc. IEEE Conf. Comp. Vis. Patt. Recogn.}, June 2019.

\bibitem{metaoptnet}
K.~Lee, S.~Maji, A.~Ravichandran, and S.~Soatto, ``Meta-learning with
  differentiable convex optimization,'' in \emph{Proc. IEEE Conf. Comp. Vis.
  Patt. Recogn.}, 2019.

\bibitem{Generative}
S.~Bartunov and D.~P. Vetrov, ``Few-shot generative modelling with generative
  matching networks,'' in \emph{Proc. Int. Conf. Artificial Intell. \& Stat.},
  2018.

\bibitem{Imaginary}
Y.~Wang, R.~B. Girshick, M.~Hebert, and B.~Hariharan, ``Low-shot learning from
  imaginary data,'' in \emph{Proc. IEEE Conf. Comp. Vis. Patt. Recogn.}, 2018.

\bibitem{SchwartzNIPS18}
E.~Schwartz, L.~Karlinsky, J.~Shtok, S.~Harary, M.~Marder, R.~S. Feris,
  A.~Kumar, R.~Giryes, and A.~M. Bronstein, ``Delta-encoder: an effective
  sample synthesis method for few-shot object recognition,'' in \emph{Proc.
  Advances in Neural Inf. Process. Syst.}, 2018.

\bibitem{Mehrotra2017}
A.~Mehrotra and A.~Dukkipati, ``Generative adversarial residual pairwise
  networks for one shot learning,'' \emph{arXiv}, vol. 1703.08033, 2017.

\bibitem{ZhangNIPS2018MetaGAN}
R.~Zhang, T.~Che, Z.~Grahahramani, Y.~Bengio, and Y.~Song, ``Metagan: An
  adversarial approach to few-shot learning,'' in \emph{Proc. Advances in
  Neural Inf. Process. Syst.}, 2018.

\bibitem{Hallucination}
H.~Zhang, J.~Zhang, and P.~Koniusz, ``Few-shot learning via saliency-guided
  hallucination of samples,'' in \emph{Proc. IEEE Conf. Comp. Vis. Patt.
  Recogn.}, June 2019.

\bibitem{shen2019learning}
W.~Shen, Z.~Shi, and J.~Sun, ``Learning from adversarial features for few-shot
  classification,'' \emph{arXiv preprint arXiv:1903.10225}, 2019.

\bibitem{li2020adversarial}
K.~Li, Y.~Zhang, K.~Li, and Y.~Fu, ``Adversarial feature hallucination networks
  for few-shot learning,'' in \emph{Proc. IEEE Conf. Comp. Vis. Patt. Recogn.},
  2020, pp. 13\,470--13\,479.

\bibitem{yang2021free}
S.~Yang, L.~Liu, and M.~Xu, ``Free lunch for few-shot learning: Distribution
  calibration,'' \emph{arXiv: Comp. Res. Repository}, 2021.

\bibitem{rodriguez2020embedding}
P.~Rodr{\'\i}guez, I.~Laradji, A.~Drouin, and A.~Lacoste, ``Embedding
  propagation: Smoother manifold for few-shot classification,'' in \emph{Proc.
  Eur. Conf. Comp. Vis.}\hskip 1em plus 0.5em minus 0.4em\relax Springer, 2020,
  pp. 121--138.

\bibitem{dhillon2019baseline}
G.~S. Dhillon, P.~Chaudhari, A.~Ravichandran, and S.~Soatto, ``A baseline for
  few-shot image classification,'' \emph{arXiv: Comp. Res. Repository}, 2019.

\bibitem{hu2020empirical}
S.~X. Hu, P.~G. Moreno, Y.~Xiao, X.~Shen, G.~Obozinski, N.~D. Lawrence, and
  A.~Damianou, ``Empirical {B}ayes transductive meta-learning with synthetic
  gradients,'' \emph{arXiv: Comp. Res. Repository}, 2020.

\bibitem{yue2020interventional}
Z.~Yue, H.~Zhang, Q.~Sun, and X.-S. Hua, ``Interventional few-shot learning,''
  \emph{arXiv preprint arXiv:2009.13000}, 2020.

\bibitem{boudiaf2020transductive}
M.~Boudiaf, Z.~I. Masud, J.~Rony, J.~Dolz, P.~Piantanida, and I.~B. Ayed,
  ``Transductive information maximization for few-shot learning,'' \emph{arXiv
  preprint arXiv:2008.11297}, 2020.

\bibitem{shyam2017attentive}
P.~Shyam, S.~Gupta, and A.~Dukkipati, ``Attentive recurrent comparators,'' in
  \emph{Proc. Int. Conf. Mach. Learn.}\hskip 1em plus 0.5em minus 0.4em\relax
  JMLR. org, 2017, pp. 3173--3181.

\bibitem{lstmmeta}
S.~Ravi and H.~Larochelle, ``Optimization as a model for few-shot learning,''
  in \emph{Proc. Int. Conf. Learn. Representations}, 2017.

\bibitem{su2020does}
J.-C. Su, S.~Maji, and B.~Hariharan, ``When does self-supervision improve
  few-shot learning?'' in \emph{Proc. Eur. Conf. Comp. Vis.}\hskip 1em plus
  0.5em minus 0.4em\relax Springer, 2020, pp. 645--666.

\bibitem{tian2020rethinking}
Y.~Tian, Y.~Wang, D.~Krishnan, J.~B. Tenenbaum, and P.~Isola, ``Rethinking
  few-shot image classification: a good embedding is all you need?''
  \emph{arXiv preprint arXiv:2003.11539}, 2020.

\bibitem{wu2020attentive}
F.~Wu, J.~S. Smith, W.~Lu, C.~Pang, and B.~Zhang, ``Attentive prototype
  few-shot learning with capsule network-based embedding,'' in \emph{Proc. Eur.
  Conf. Comp. Vis.}\hskip 1em plus 0.5em minus 0.4em\relax Springer, 2020, pp.
  237--253.

\bibitem{SNAIL}
N.~Mishra, M.~Rohaninejad, X.~Chen, and P.~Abbeel, ``Snail: A simple neural
  attentive meta-learner,'' in \emph{Proc. Int. Conf. Learn. Representations},
  2018.

\bibitem{zhang2019canet}
C.~Zhang, G.~Lin, F.~Liu, R.~Yao, and C.~Shen, ``Canet: Class-agnostic
  segmentation networks with iterative refinement and attentive few-shot
  learning,'' in \emph{Proc. IEEE Conf. Comp. Vis. Patt. Recogn.}, 2019, pp.
  5217--5226.

\bibitem{zhang2019pyramid}
C.~Zhang, G.~Lin, F.~Liu, J.~Guo, Q.~Wu, and R.~Yao, ``Pyramid graph networks
  with connection attentions for region-based one-shot semantic segmentation,''
  in \emph{Proc. IEEE Int. Conf. Comp. Vis.}, 2019, pp. 9587--9595.

\bibitem{liu2020crnet}
W.~Liu, C.~Zhang, G.~Lin, and F.~Liu, ``{CRNet}: Cross-reference networks for
  few-shot segmentation,'' \emph{arXiv: Comp. Res. Repository}, 2020.

\bibitem{yang2020contexttransformer}
Z.~Yang, Y.~Wang, X.~Chen, J.~Liu, and Y.~Qiao, ``Context-transformer: Tackling
  object confusion for few-shot detection,'' \emph{arXiv: Comp. Res.
  Repository}, 2020.

\bibitem{kusner2015word}
M.~Kusner, Y.~Sun, N.~Kolkin, and K.~Weinberger, ``From word embeddings to
  document distances,'' in \emph{Proc. Int. Conf. Mach. Learn.}, 2015, pp.
  957--966.

\bibitem{wang2014new}
C.~Wang and S.~Chan, ``A new hand gesture recognition algorithm based on joint
  color-depth superpixel earth mover's distance,'' in \emph{Proc. Int. Workshop
  Cognitive Information Processing}.\hskip 1em plus 0.5em minus 0.4em\relax
  IEEE, 2014, pp. 1--6.

\bibitem{nikolentzos2017matching}
G.~Nikolentzos, P.~Meladianos, and M.~Vazirgiannis, ``Matching node embeddings
  for graph similarity,'' in \emph{Proc. {AAAI} Conf. Artificial Intell.},
  2017.

\bibitem{schulter2017deep}
S.~Schulter, P.~Vernaza, W.~Choi, and M.~Chandraker, ``Deep network flow for
  multi-object tracking,'' in \emph{Proc. IEEE Conf. Comp. Vis. Patt. Recogn.},
  2017, pp. 6951--6960.

\bibitem{zhao2008differential}
Q.~Zhao, Z.~Yang, and H.~Tao, ``Differential earth mover's distance with its
  applications to visual tracking,'' \emph{{IEEE} Trans. Pattern Anal. Mach.
  Intell.}, vol.~32, no.~2, pp. 274--287, 2008.

\bibitem{li2013tensor}
P.~Li, ``Tensor-sift based earth mover’s distance for contour tracking,''
  \emph{J. Mathematical Imaging \& Vision}, vol.~46, no.~1, pp. 44--65, 2013.

\bibitem{gould2016differentiating}
S.~Gould, B.~Fernando, A.~Cherian, P.~Anderson, R.~S. Cruz, and E.~Guo, ``On
  differentiating parameterized argmin and argmax problems with application to
  bi-level optimization,'' \emph{arXiv preprint arXiv:1607.05447}, 2016.

\bibitem{agrawal2019differentiating}
A.~Agrawal, S.~Barratt, S.~Boyd, E.~Busseti, and W.~M. Moursi,
  ``Differentiating through a conic program,'' \emph{arXiv preprint
  arXiv:1904.09043}, 2019.

\bibitem{amos2017optnet}
B.~Amos and J.~Z. Kolter, ``{O}pt{N}et: Differentiable optimization as a layer
  in neural networks,'' in \emph{Proc. Int. Conf. Mach. Learn.}, vol.~70, 2017,
  pp. 136--145.

\bibitem{agrawal2019differentiable}
A.~Agrawal, B.~Amos, S.~Barratt, S.~Boyd, S.~Diamond, and J.~Z. Kolter,
  ``Differentiable convex optimization layers,'' in \emph{Proc. Advances in
  Neural Inf. Process. Syst.}, 2019, pp. 9558--9570.

\bibitem{vlastelica2019differentiation}
M.~Vlastelica, A.~Paulus, V.~Musil, G.~Martius, and M.~Rol{\'\i}nek,
  ``Differentiation of blackbox combinatorial solvers,'' \emph{arXiv preprint
  arXiv:1912.02175}, 2019.

\bibitem{rolinek2020deep}
M.~Rol{\'\i}nek, P.~Swoboda, D.~Zietlow, A.~Paulus, V.~Musil, and G.~Martius,
  ``Deep graph matching via blackbox differentiation of combinatorial
  solvers,'' \emph{arXiv preprint arXiv:2003.11657}, 2020.

\bibitem{fcn}
E.~Shelhamer, J.~Long, and T.~Darrell, ``Fully convolutional networks for
  semantic segmentation,'' \emph{{IEEE} Trans. Pattern Anal. Mach. Intell.},
  vol.~39, no.~4, pp. 640--651, 2017.

\bibitem{weightnorm}
T.~Salimans and D.~P. Kingma, ``Weight normalization: A simple
  reparameterization to accelerate training of deep neural networks,'' in
  \emph{Proc. Advances in Neural Inf. Process. Syst.}, 2016, pp. 901--909.

\bibitem{cub}
C.~Wah, S.~Branson, P.~Welinder, P.~Perona, and S.~Belongie, ``The caltech-ucsd
  birds-200-2011 dataset,'' 2011.

\bibitem{bertinetto2018meta}
L.~Bertinetto, J.~F. Henriques, P.~H. Torr, and A.~Vedaldi, ``Meta-learning
  with differentiable closed-form solvers,'' \emph{arXiv preprint
  arXiv:1805.08136}, 2018.

\bibitem{imagenet}
O.~Russakovsky, J.~Deng, H.~Su, J.~Krause, S.~Satheesh, S.~Ma, Z.~Huang,
  A.~Karpathy, A.~Khosla, M.~S. Bernstein, A.~C. Berg, and F.~Li, ``Imagenet
  large scale visual recognition challenge,'' \emph{Int. J. Comput. Vision},
  vol. 115, no.~3, pp. 211--252, 2015.

\bibitem{CIFAR100}
A.~Krizhevsky, ``Learning multiple layers of features from tiny images,''
  \emph{University of Toronto}, 2009.

\bibitem{OpenCV}
G.~Bradski and A.~Kaehler, \emph{Learning OpenCV: Computer vision with the
  OpenCV library}.\hskip 1em plus 0.5em minus 0.4em\relax " O'Reilly Media,
  Inc.", 2008.

\bibitem{kmeans}
\BIBentryALTinterwordspacing
X.~Jin and J.~Han, \emph{K-Means Clustering}.\hskip 1em plus 0.5em minus
  0.4em\relax Boston, MA: Springer US, 2010, pp. 563--564. [Online]. Available:
  \url{https://doi.org/10.1007/978-0-387-30164-8_425}
\BIBentrySTDinterwordspacing

\bibitem{meanshift}
D.~Comaniciu and P.~Meer, ``Mean shift: A robust approach toward feature space
  analysis,'' \emph{IEEE Trans. Pattern Anal. Mach. Intell.}, vol.~24, no.~5,
  pp. 603--619, May 2002.

\bibitem{scikit-learn}
F.~Pedregosa, G.~Varoquaux, A.~Gramfort, V.~Michel, B.~Thirion, O.~Grisel,
  M.~Blondel, P.~Prettenhofer, R.~Weiss, V.~Dubourg, J.~Vanderplas, A.~Passos,
  D.~Cournapeau, M.~Brucher, M.~Perrot, and E.~Duchesnay, ``Scikit-learn:
  Machine learning in {P}ython,'' \emph{J. Machine Learning Research}, vol.~12,
  pp. 2825--2830, 2011.

\bibitem{ECM}
A.~Ravichandran, R.~Bhotika, and S.~Soatto, ``Few-shot learning with embedded
  class models and shot-free meta training,'' \emph{arXiv preprint
  arXiv:1905.04398}, 2019.

\bibitem{tpn}
Y.~Liu, J.~Lee, M.~Park, S.~Kim, E.~Yang, S.~J. Hwang, and Y.~Yang, ``Learning
  to propagate labels: Transductive propagation network for few-shot
  learning,'' \emph{arXiv preprint arXiv:1805.10002}, 2018.

\bibitem{mangla2020charting}
P.~Mangla, N.~Kumari, A.~Sinha, M.~Singh, B.~Krishnamurthy, and V.~N.
  Balasubramanian, ``Charting the right manifold: Manifold mixup for few-shot
  learning,'' in \emph{The IEEE Winter Conference on Applications of Computer
  Vision}, 2020, pp. 2218--2227.

\bibitem{liu2020ensemble}
Y.~Liu, B.~Schiele, and Q.~Sun, ``An ensemble of epoch-wise empirical bayes for
  few-shot learning,'' in \emph{Proc. Eur. Conf. Comp. Vis.}\hskip 1em plus
  0.5em minus 0.4em\relax Springer, 2020, pp. 404--421.

\bibitem{moco}
K.~He, H.~Fan, Y.~Wu, S.~Xie, and R.~Girshick, ``Momentum contrast for
  unsupervised visual representation learning,'' in \emph{Proc. IEEE Conf.
  Comp. Vis. Patt. Recogn.}, 2020.

\bibitem{cuturi2013sinkhorn}
M.~Cuturi, ``Sinkhorn distances: Lightspeed computation of optimal transport,''
  \emph{Proc. Advances in Neural Inf. Process. Syst.}, vol.~26, pp. 2292--2300,
  2013.

\bibitem{musgrave2020metric}
K.~Musgrave, S.~Belongie, and S.-N. Lim, ``A metric learning reality check,''
  in \emph{Proc. Eur. Conf. Comp. Vis.}\hskip 1em plus 0.5em minus 0.4em\relax
  Springer, 2020, pp. 681--699.

\bibitem{zhai2018classification}
A.~Zhai and H.-Y. Wu, ``Classification is a strong baseline for deep metric
  learning,'' \emph{arXiv preprint arXiv:1811.12649}, 2018.

\bibitem{hadsell2006dimensionality}
R.~Hadsell, S.~Chopra, and Y.~LeCun, ``Dimensionality reduction by learning an
  invariant mapping,'' in \emph{Proc. IEEE Conf. Comp. Vis. Patt. Recogn.},
  2006, pp. 1735--1742.

\bibitem{qian2019softtriple}
Q.~Qian, L.~Shang, B.~Sun, J.~Hu, H.~Li, and R.~Jin, ``Softtriple loss: Deep
  metric learning without triplet sampling,'' in \emph{Proc. IEEE Int. Conf.
  Comp. Vis.}, 2019, pp. 6450--6458.

\bibitem{yuan2019signal}
T.~Yuan, W.~Deng, J.~Tang, Y.~Tang, and B.~Chen, ``Signal-to-noise ratio: A
  robust distance metric for deep metric learning,'' in \emph{Proc. IEEE Conf.
  Comp. Vis. Patt. Recogn.}, 2019, pp. 4815--4824.

\bibitem{deng2019arcface}
J.~Deng, J.~Guo, N.~Xue, and S.~Zafeiriou, ``Arcface: Additive angular margin
  loss for deep face recognition,'' in \emph{Proc. IEEE Conf. Comp. Vis. Patt.
  Recogn.}, 2019, pp. 4690--4699.

\bibitem{wang2019multi}
X.~Wang, X.~Han, W.~Huang, D.~Dong, and M.~R. Scott, ``Multi-similarity loss
  with general pair weighting for deep metric learning,'' in \emph{Proc. IEEE
  Conf. Comp. Vis. Patt. Recogn.}, 2019, pp. 5022--5030.

\end{thebibliography}
